\documentclass[letterpaper]{article} % DO NOT CHANGE THIS
\usepackage{aaai23}  % DO NOT CHANGE THIS
\usepackage{times}  % DO NOT CHANGE THIS
\usepackage{helvet}  % DO NOT CHANGE THIS
\usepackage{courier}  % DO NOT CHANGE THIS
\usepackage[hyphens]{url}  % DO NOT CHANGE THIS
\usepackage{graphicx} % DO NOT CHANGE THIS
\usepackage{natbib}  % DO NOT CHANGE THIS AND DO NOT ADD ANY OPTIONS TO IT
\usepackage{caption} % DO NOT CHANGE THIS AND DO NOT ADD ANY OPTIONS TO IT
\usepackage{algorithm}
\usepackage{algorithmic}
\usepackage{xspace}
\usepackage[dvipsnames]{xcolor}
\usepackage{paralist}
\usepackage{bm}
\usepackage{bbm}
\usepackage{amssymb}
\usepackage{amsfonts} 
\usepackage{amsmath} 
\usepackage{caption}
\usepackage{subcaption}
\usepackage{booktabs}
\usepackage{etoolbox}
\usepackage{tikz}
\usepackage{pifont}
\usepackage{makecell}
\usepackage{enumitem}
\usepackage{mathtools}
\usepackage{circledsteps}

\usepackage{newfloat}
\usepackage{listings}
\DeclareCaptionStyle{ruled}{labelfont=normalfont,labelsep=colon,strut=off} % DO NOT CHANGE THIS
\floatstyle{ruled}
\newfloat{listing}{tb}{lst}{}
\floatname{listing}{Listing}

\newcommand{\method}{{ADMoE}\xspace}
\newcommand{\prob}{{MNLAD}\xspace}
\newcommand{\am}{{ADMoE}\xspace}

\newcommand{\cbit}{\begin{compactitem}}
	\newcommand{\ceit}{\end{compactitem}}
\newcommand{\cben}{\begin{compactenum}}
	\newcommand{\ceen}{\end{compactenum}}

\newtheorem{problem}{Problem}

\newrobustcmd*{\mysquare}[1]{\tikz{\filldraw[draw=#1,fill=#1] (0,0)
rectangle (0.18cm,0.18cm);}}

\newrobustcmd*{\mycircle}[1]{\tikz{\filldraw[draw=#1,fill=#1] (0,0) circle [radius=0.1cm];}}

\newrobustcmd*{\mytriangle}[1]{\tikz{\filldraw[draw=#1,fill=#1] (0,0) --
(0.2cm,0) -- (0.1cm,0.2cm);}}

\definecolor{dodgerblue}{rgb}{0.12,0.565,1}
\definecolor{moeblue}{rgb}{0.792,0.938,0.99}
\definecolor{moegreen}{rgb}{0.742,0.859,0.683}

\title{\method: Anomaly Detection with Mixture-of-Experts from Noisy Labels}

\author{
    Yue Zhao$^1$\thanks{The project is primarily done at Microsoft Research.}, Guoqing Zheng$^2$, Subhabrata Mukherjee$^2$, Robert McCann$^3$, Ahmed Awadallah$^2$\\
}
\affiliations{
    $^1$Carnegie Mellon University, $^2$Microsoft Research, $^3$Microsoft\\
    zhaoy@cmu.edu, \{guoqing.zheng,submukhe,robert.mccann,hassanam\}@microsoft.com

}

\usepackage{bibentry}
\begin{document}

\maketitle

\begin{abstract}

Existing works on anomaly detection (AD) rely on clean labels from human annotators that are expensive to acquire in practice.  In this work, we propose a method to leverage weak/noisy labels (e.g., risk scores generated by machine rules for detecting malware) that are cheaper to obtain for anomaly detection. Specifically, we propose \method, \textit{the first framework for anomaly detection algorithms to learn from noisy labels}. In a nutshell, \method leverages Mixture-of-experts (MoE) architecture to encourage \textit{specialized} and \textit{scalable} learning from multiple noisy sources. It captures the similarities among noisy labels by sharing most model parameters, while encouraging specialization by building ``expert'' sub-networks. To further juice out the signals from noisy labels, \method uses them as input features to facilitate expert learning. Extensive results on eight datasets (including a proprietary enterprise security dataset) demonstrate the effectiveness of \method, where it brings up to 34\% performance improvement over not using it. Also, it outperforms a total of 13 leading baselines with equivalent network parameters and FLOPS. Notably, \method is model-agnostic to enable any neural network-based detection methods to handle noisy labels, where we showcase its results on both multiple-layer perceptron (MLP) and leading AD method DeepSAD.
\end{abstract}

\section{Introduction}
\label{sec:intro}
Anomaly detection (AD), also known as outlier detection, is a crucial learning task with many real-world applications, including malware detection \cite{nguyen2019diot}, anti-money laundering \citep{lee2020autoaudit}, rare-disease detection \citep{li2018semi} and so on. 
Although there are numerous detection algorithms \citep{Aggarwal2013a,pang2021deep,zhao2021automatic,liu2022bond}, 
existing AD methods assume the availability of (partial) labels that are \textit{clean} (i.e. without noise), and cannot learn from weak/noisy labels\footnote{We use the terms \textit{noisy} and \textit{weak} interchangeably.}. 

Simply treating noisy labels as (pseudo) clean labels leads to biased and degraded models \cite{song2022learning}. Over the years, researchers have developed algorithms for classification and regression tasks to learn from noisy sources \cite{rodrigues2018deep,guan2018said,wei2022deep}, which has shown great success.
However, these methods are not tailored for AD with extreme data imbalance, and existing AD methods cannot learn from (multiple) noisy sources.

Why is it important to leverage noisy labels in AD applications? 
Taking malware detection as an example, it is impossible to get a large number of clean labels due to the data sensitivity and the cost of annotation.
However, often there exists
a large number of 
% (up to hundreds) 
weak/noisy historical security rules designed for detecting malware from different perspectives, e.g., unauthorized network access and suspicious file movement, which have not been used in AD yet.
Though not as perfect as human annotations, they
% these noisy sources  
% generated by human experts or weak machine detectors 
are valuable as they encode prior knowledge from past detection experiences.
Also, although each noisy source may be insufficient for difficult AD tasks, learning them jointly may build competitive models as they tend to \textit{complement} each other.

\begin{figure} [!t]
    \centering
	\subfloat[\scriptsize Comparison with leading AD methods]{%
		\includegraphics[clip,width=0.485\columnwidth]{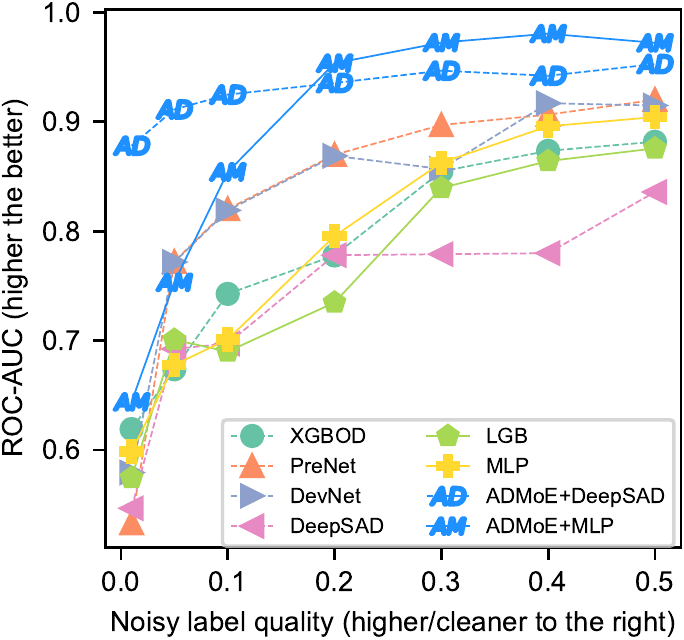}%
	\label{fig:mlp_demo}
	}
% 	\vspace{0.1in}
	\subfloat[\scriptsize Comp. w/ noisy learning methods] {%
		\includegraphics[clip,width=0.485\columnwidth]{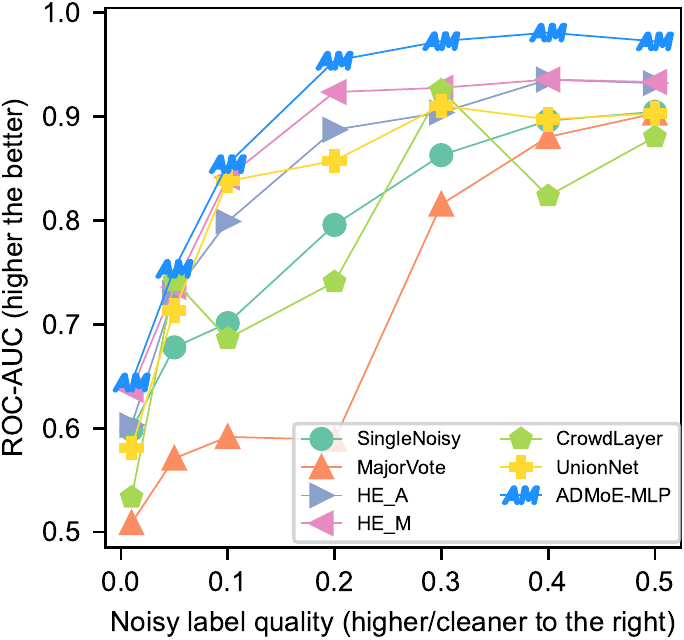}%
		\label{fig:perf_highlight}
	}
	\vspace{-0.1in}
	\caption{Performance (ROC-AUC) comparison on \texttt{Yelp} (see results on all datasets in \S \ref{exp:leading_AD} and \ref{exp:leading_nl}), where \method outperforms two groups of baselines: (a) SOTA AD methods; (b) leading classification methods for learning from multiple noisy sources. \method enhanced DeepSAD and MLP are denoted as \textcolor{dodgerblue}{\textbf{\textit{AD}}} and \textcolor{dodgerblue}{\textbf{\textit{AM}}}.}
	\label{fig:intro_performance}
	\vspace{-0.25in}
\end{figure}

In this work, we propose \method, (to our knowledge) the \textit{first weakly-supervised approach} for enabling anomaly detection algorithms to learn from 
 \textit{multiple sets of noisy labels}. 
In a nutshell, \method enhances existing neural-network-based AD algorithms by Mixture-of-experts (MoE) network(s) \cite{jacobs1991adaptive,MoE}, which has a learnable gating function to activate different sub-networks (experts) based on the \textit{incoming samples} and \textit{their noisy labels}. In this way, the proposed \method can jointly learn from multiple sets of noisy labels with the majority of parameters shared, while providing specialization and scalability via experts. Unlike existing noisy label learning approaches, \method does not require 
% a one-to-one 
explicit mapping from noisy labels to network parameters, providing better scalability and flexibility. To encourage \method to develop specialization based on the noisy sources, 
we use noisy labels as (part of the) input features with learnable embeddings to make the gating function aware of them.

\textbf{Key Results}.
Fig. \ref{fig:intro_performance}
shows that a multiple layer perception (MLP) \cite{rosenblatt1958perceptron} enhanced by \method
can largely outperform both (\ref{fig:mlp_demo}) leading AD algorithms as well as (\ref{fig:perf_highlight}) noisy-label learning methods for classification. Note \method is not strictly another detection algorithm, but \textit{a general framework} to empower any neural-based AD methods to leverage multiple sets of weak labels. 
% Fig. \ref{fig:ours} shows how to apply it to an MLP.
\S \ref{sec:exp} shows extensive results on more datasets, and the  improvement in enhancing more complex DeepSAD \cite{ruff2019deep} with \method.

In summary, the key contributions of this work include:
\cbit
\item \textbf{Problem formulation, baselines, and datasets}. We formally define the crucial problem of using {\em \underline{m}ultiple sets of \underline{n}oisy \underline{l}abels for \underline{AD}} (\prob), and release the first batch of baselines and datasets for future research\footnote{See code and appendix: \url{https://github.com/microsoft/admoe}}.
\item \textbf{The first AD framework for learning from multiple noisy sources}. The proposed \method is a novel method with Mixture-of-experts (MoE) architecture to achieve specialized and scalable learning for \prob.
\item \textbf{Model-agnostic design}. \method enhances any neural-network-based AD methods, and we show its effectiveness on MLP and state-of-the-art (SOTA) DeepSAD.
\item \textbf{Effectiveness and real-world deployment}. 
We demonstrate \method's SOTA performance on seven benchmark datasets and a proprietary enterprise security application, in comparison with two groups of leading baselines (13 in total). It brings on average 14\% and up to 34\% improvement over not using it, with the equivalent number of learnable parameters and FLOPs as baselines. 
\ceit

% \vspace{-0.1in}

\section{Related Work}
\label{sec:related}
\subsection{Weakly-supervised Anomaly Detection}
There exists some literature on weakly-supervised AD, and most of them fall under the ``incomplete supervision'' category. These semi-supervised methods assume access to a small set of clean labels and it is unclear how to extend them for multi-set noisy labels. Representative work includes XGBOD \cite{zhao2018xgbod}, DeepSAD \cite{ruff2019deep}, DevNet \cite{pang2019devnet}, PreNet \cite{pang2019deep}; see Appx. \ref{appx:semiad} for more details. We use these leading methods (that work with one set of labels) as baselines in \S \ref{exp:leading_AD}, and demonstrate \method's performance gain in leveraging multiple sets of noisy labels.
\vspace{-0.1in}

\subsection{Learning from Single Set of Noisy Labels}
There have been rich literature on learning from a single set of noisy labels, including learning a label corruption/transition matrix \cite{patrini2017making}, correcting labels via meta-learning \cite{zheng2021meta}, and building robust training mechanisms like co-teaching \cite{han2018co}, co-teaching+ \cite{yu2019does}, and JoCoR \cite{wei2020combating}. More details can be found in a recent survey \cite{han2020survey}. These algorithms are primarily for a single set of noisy labels, and are not designed for AD tasks.

\vspace{-0.15in}

\subsection{Learning from Multiple Noisy Sources}
\label{sec:related_b2}
\setlength\tabcolsep{1.5 pt}
\newcommand{\cmark}{\ding{51}}%
\newcommand{\xmark}{\ding{55}}%

\begin{table}[!ht]
\vspace{-0.1in}
\centering
% 	\foot\xmarktesize
\vspace{-0.1in}
\scalebox{0.66}{
\begin{tabular}{l|cc|cc|cc|c}
\toprule
\textbf{Category}      & \textbf{SingleNoisy} & \textbf{LabelVote} & \textbf{HE\_A} & \textbf{HE\_M} & \textbf{CrowdLayer}  & \textbf{UnionNet} & \textcolor{dodgerblue}{\textbf{\method}} \\
\midrule
\textbf{Multi-source} & \xmark    & \cmark            & \cmark                & \cmark                & \cmark                 & \cmark                              & \textcolor{dodgerblue}{\cmark}            \\
\textbf{Single-model} & \cmark    & \cmark            & \xmark                & \xmark                & \cmark                 & \cmark                              & \textcolor{dodgerblue}{\cmark}            \\
\textbf{End-to-end}    & \cmark & \xmark                 & \xmark                & \xmark                & \cmark                 & \cmark                      & \textcolor{dodgerblue}{\cmark}           \\
\textbf{Scalability} &    High  & High               & Low               & Low               & Med                 & Med                & \textcolor{dodgerblue}{High}       \\
\bottomrule
\end{tabular}
}
	\caption{Baselines and \method for comparison with categorization by (first row) whether it uses multiple sets of weak labels, (second row) whether it only trains a single model, (third row) whether the training process is end-to-end and (the last row) whether it is scalable with regard to many sets of weak labels. \method is an end-to-end, scalable paradigm. } % title of Table
	\label{table:baselines} % is used to refer this table in the text
	\vspace{-0.15in}
\end{table}
\setlength\tabcolsep{6 pt}

% \scalebox{0.65}{
% \begin{tabular}{l|c|cc|cccc|c}
% \toprule
% \textbf{Category}      & \textbf{LabelVote} & \textbf{HE\_A} & \textbf{HE\_M} & \textbf{CrowdLayer} & \textbf{DoctorNet} & \textbf{WDN} & \textbf{UnionNet} & \textcolor{dodgerblue}{\textbf{\method}} \\
% \midrule
% \textbf{SingleModel} & \cmark                & \xmark                & \xmark                & \cmark                 & \cmark                & \cmark          & \cmark               & \textcolor{dodgerblue}{\cmark}            \\
% \textbf{End-to-end}    & \xmark                 & \xmark                & \xmark                & \cmark                 & \cmark                & \cmark          & \cmark               & \textcolor{dodgerblue}{\cmark}           \\
% \textbf{Scalability}     & High               & Low               & Low               & Med                 & Med                & Med          & Med               & \textcolor{dodgerblue}{High}       \\
% \bottomrule
% \end{tabular}
% }

\prob falls under \textit{weakly supervised ML}  \cite{zhou2018brief}, where it deals with \textit{multiple} sets of \textit{inaccurate/noisy} labels. Naturally, one can aggregate noisy labels to generate a ``corrected'' label set \cite{zheng2021meta}, while it may be challenging for AD with extreme data imbalance.
% However, AD problem faces extreme data imbalance (i.e., anomalies are way rare than normal samples), and it is hard to correct noisy labels directly.
Other than label correction, one may take ensembling to train multiple independent AD models for combination, e.g., one model per set of noisy labels, which however faces scalability issues while dealing with many sets of labels. What is worse, independently trained models fail to explore the interaction among noisy labels. Differently, end-to-end noisy label learning methods, including Crowd Layer \cite{rodrigues2018deep}, DoctorNet \cite{guan2018said}, and UnionNet \cite{wei2022deep}, can directly learn from multiple sets of noisy labels and map each set of noisy labels to part of the network (e.g., transition matrix), encouraging the model to learn knowledge from all noisy labels collectively. Although they yield great performance in crowd-sourcing scenarios with a small number of annotators, they do not scale in \prob with many sets of ``cheap'' noisy labels 
% automatically generated by rules or machines 
due to this explicit one-to-one mapping. Also, each set of noisy AD labels may be only good at certain anomalies as a biased annotator. Consequently, 
% , where in crowd-sourcing scenarios they are more globally accurate. To that end, 
explicit one-to-one mapping (e.g., transition matrix) from a single set of labels to a network in existing classification works 
% is a strong constraint and 
is not ideal for \prob.
\method lifts this constraint to allow many-to-many mapping, improving model scalability and robustness for \prob.

We summarize the 
% these classical and leading 
methods for learning from multiple sources of noisy labels
% in crowd-sourcing 
in Table \ref{table:baselines} categorized as: 
% \cbit
% \item 
\textbf{(1) SingleNoisy} trains an AD model using only one set of weak labels, which sets the lower bound of all baselines.
% \item 
\textbf{(2) LabelVote} trains an AD model based on the consensus of weak labels via majority vote.
% \item 
\textbf{(3) HyperEnsemble} \cite{wenzel2020hyperparameter} trains an individual model for each set of noisy labels (i.e., $k$ models for $k$ sets of labels), and combines their anomaly scores by averaging (i.e., \textbf{HE\_A}) and maximizing (i.e., \textbf{HE\_M}). 
% \item 
\textbf{(4) CrowdLayer} \cite{rodrigues2018deep} tries to reconstruct the input weak labels during training.
\textbf{(5) UnionNet} \cite{wei2022deep} learns a transition matrix for all weak labels together.
% \ceit
Note that they are primarily for classification and not tailored for anomaly detection. Nonetheless, we adapt the methods in Table \ref{table:baselines} as baselines.
In \S \ref{exp:leading_nl}, we show \method outperforms all these methods.

\vspace{-0.1in}

% \section{Anomaly Detection from Multiple Noisy Sources}
\section{AD from Multiple Sets of Noisy Labels}
\label{sec:method}
In \S \ref{subsuc:prob}, we formally present the problem of AD with multiple sets of noisy labels, followed by the discussion on why multiple noisy sources help AD 
% and the limitations of existing non-AD methods 
in \S \ref{subsec:why_how_multiple}. Motivated by above, we describe the proposed \method framework 
% \method
% and its key merits 
in \S \ref{subsec:algorithm}. 
\vspace{-0.05in}

\subsection{Problem Statement}
\label{subsuc:prob}
We consider the problem of anomaly detection with multiple sets of weak/noisy labels. We refer to this problem as \prob, an acronym for using \underline{m}ultiple sets of \underline{n}oisy \underline{l}abels for \underline{a}nomaly \underline{d}etection. 
% Formally, 
We present the problem definition here.

\begin{problem}
% [HP Tuning for Unsupervised Outlier Detection (\prob)]
[\prob]
\textit{\em Given} an anomaly detection task with input feature $\mathbf{X} \in \mathbb{R}^{n\times d}$ (e.g., $n$ samples and $d$ features)
% along with an optional small set  ($n_c \ll n$) of clean labels $\mathbf{y^c} \in \mathbb{R}^{n_c}$  
and $t$ sets of noisy/weak labels $\mathcal{Y}_w =  \{\mathbf{y}_{w,1},\dots,\mathbf{y}_{w,t}\}$ (each in $\mathbb{R}^{n}$), 
build a detection model $M$ to leverage all information to achieve the best performance.
\label{problem_def}
\end{problem}
\vspace{-0.15in}

Existing 
% semi- and fully-supervised 
AD methods \cite{Aggarwal2013a,pang2021deep,han2022adbench} can (at best) 
% leverage a small set of (optional) clean labels $\mathbf{y^c}$, and
% /or may 
treat one set of noisy labels from $\mathcal{Y}^w$ as (pseudo) clean labels to train a model. 
% However, 
None of them leverages multiple sets of noisy labels $\mathcal{Y}^w$ collectively.
\vspace{-0.15in}

\subsection{Why and How Do Multiple Sets of Weak Labels Help in Anomaly Detection?}
\label{subsec:why_how_multiple}
\textbf{Benefits of joint learning in \prob}. 
% It is well known that 
AD benefits from model combination and ensemble learning
% and joint learning 
from diverse base models \cite{aggarwal2017outlier,zhao2019lscp,ding2022hyperparameter}, 
and the improvement is expected when base models make \textit{complementary} errors \cite{zimek2014ensembles,aggarwal2017outlier}. 
Multiple sets of noisy labels are natural sources 
% for constructing 
for ensembling with built-in diversity, as they reflect distinct detection aspects and historical knowledge.
Appx. \ref{appx:prob_examples} Fig. \ref{fig:perf_ensemble} shows that 
% HyperEnsemble  \citep{wenzel2020hyperparameter} that 
averaging the outputs from multiple AD models (each trained on one set of noisy labels) 
% ; \mysquare{red}
leads to better results than training each model independently; 
% with noisy labels;
% by each set of weak labels independently 
% (\mytriangle{blue}); 
this observation holds true for both deep (neural) AD models (e.g., PreNet in Appx. Fig.  \ref{fig:perf_ensemble_prenet}) and shallow  models (e.g., XGBOD in Appx. Fig. \ref{fig:perf_ensemble_xgbod}).
% (8) simple average the performance (pick one model) (9): train $k$ models and average their scores (and/or by variation)
This example justifies \textit{the benefit of learning from multiple sets of noisy labels};
% Note that 
% Even this 
even simple averaging in \prob %of the models by noisy labels
% (i.e., HyperEnsemble) 
can already ``touch'' the performance upper bound (i.e., training a model using all clean labels).

\begin{figure*} [!t]
    \centering
	\subfloat[Crowd Layer (Rodrigues 2018)] {%
			\includegraphics[clip,width=0.55\columnwidth]{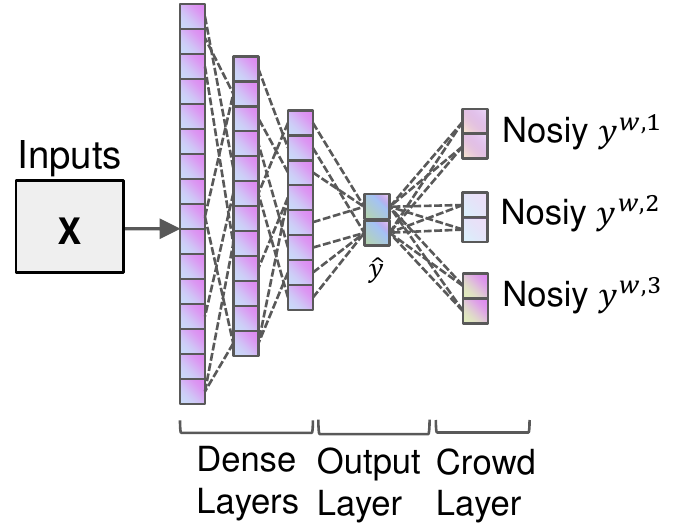}%
		\label{fig:crowdlayer}
	}
	\hspace{0.02in}
	\subfloat[Union Net \cite{wei2022deep}]{%
		\includegraphics[clip,width=0.63\columnwidth]{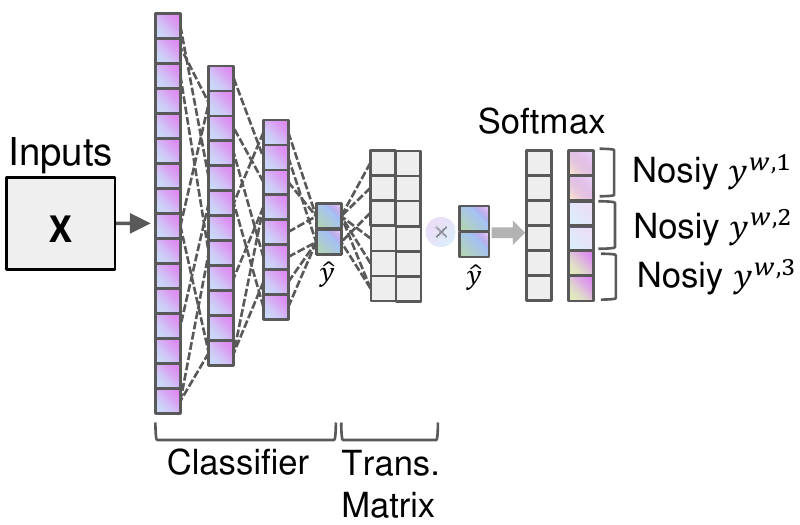}%
	\label{fig:unionnet}
	}
	\hspace{0.02in}
	\subfloat[\method (ours) leverages MoE layers for specialization] {%
		\includegraphics[clip,width=0.87\columnwidth]{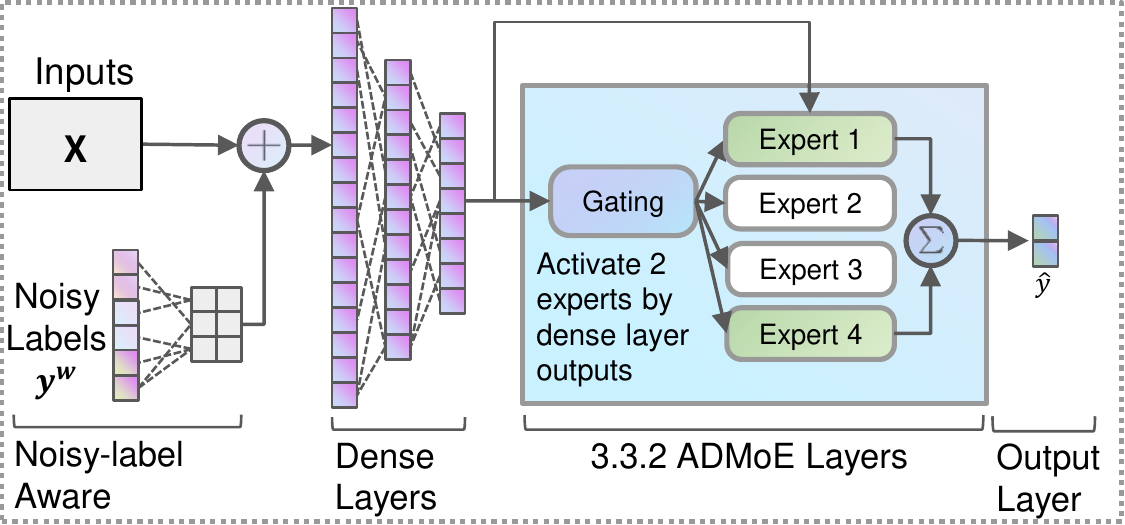}%
		\label{fig:ours}
	}
	\vspace{-0.1in}
	\caption{
	A toy example of a 3-layer MLP for AD; 3 sets of noisy labels $\mathbf{y^w}=\{\mathbf{y}^{w,1}, \mathbf{y}^{w,2}, \mathbf{y}^{w,3}\}$ are assumed. Existing methods (\textit{a} and \textit{b}) 
% 	focus on learning and recovering 
	learn to recover 
	noisy labels explicitly, while \method (\S 3.3.2) (\textit{c}) uses an MoE architecture with  
 % \method layer for
% 	two merits: 
	% implicitly
 noisy-label-aware expert activation to learn specialization from noisy sources without explicit label mapping.
 In the example, we add an \method layer (\mysquare{moeblue}) between the dense layers and the output layer; only the top two experts (\mysquare{moegreen}) are activated for 
 % each input example.
 input samples.
% 	by the gating.
	}
	\vspace{-0.25in}
	\label{fig:overall}
\end{figure*}

\subsection{\method: Specialized and Scalable Anomaly Detection with Multiple Sets of Noisy Labels}
\label{subsec:algorithm}

\textbf{Motivation}. After reviewing gaps and opportunities in existing works, we argue the ideal design for \prob should fulfill the following requirements:  (\textit{i}) encourage \textit{specialization} from different noisy sources but also explore their \textit{similarity}
% (\textit{iii}) \textit{flexibility} to not limit to one-to-one mapping from noisy labels to part of the network)
(\textit{ii}) \textit{scalability} to handle an increasing number of noisy sources
and (\textit{iii}) \textit{generality} to apply to various AD methods.

\noindent \textbf{Overview of \method}. In this work, (for the first time) we adapt Mixture-of-experts (MoE) architecture/layer \cite{jacobs1991adaptive,MoE} (see preliminary in \S 3.3.1) to AD algorithm design for \prob. 
Specifically, we propose model-agnostic \method\footnote{Throughout the paper, we slightly abuse the term \method to refer to both our overall framework and the proposed MoE layer.}
% framework
% (i.e., \underline{j}oint \underline{l}earning for \underline{a}nomaly \underline{d}etection) 
to enhance any neural network-based AD method via: 
(\textit{i}) mixture-of-experts (MoE) layers to do specialized and scalable learning from noisy labels 
% \S 3.3.2 
and (\textit{ii}) noisy-label aware expert activation by using noisy labels as (part of the) input with learnable embedding to facilitate specialization.
% encourage specialization
% in \S 3.3.3. 
Refer to Fig. \ref{fig:ours} for an illustration of applying \method to a simple MLP, where 
we add an \method layer between the dense layers and output layer for \prob,  
% (left) 
and use noisy labels directly as input with learnable embedding to help \method to better specialize.
% (right) 
Other neural-network-based AD methods can follow the same procedure to be enhanced by \method (i.e., inserting \method layers before the output layer and using noisy labels as input).
In the following subsections, we give a short background of MoE and then present the design
% and rationale 
of \method.
% as follows.
\vspace{-0.05in}

\subsubsection{3.3.1~Preliminary on Mixture-of-experts (MoE) Architecture.} 
% MoE architecture/layer \cite{jacobs1991adaptive,MoE}
The original MoE \cite{jacobs1991adaptive} is designed as a dynamic learning paradigm to allow different parts (i.e., experts) of a network to specialize for different samples. More recent (sparsely-gated) MoE \cite{MoE} has been shown to improve model scalability for natural language processing (NLP) tasks, where 
% the model can be extremely large
models can have billions of parameters \cite{du2022glam}. The key difference between the original MoE and the sparse MoE is the latter builds more experts (sub-networks) for a large model but only activates a few of them for scalability and efficiency.

Basically, MoE splits specific layer(s) of a (large) neural network into $m$ small ``experts'' (i.e., sub-networks). It uses top-$k$ gating to activate $k$ experts ($k < m$ or $k \ll m$ for sparse MoE) for each input sample computation as opposed to using the entire network as standard dense models. More specifically, MoE uses a differentiable gating function $G(\cdot)$ to calculate the activation weights of each expert. It then aggregates the weighted outputs of the top-$k$ experts with the highest activation weights as the MoE layer output. For example, the expert weights $\bm{\beta}^j \in \mathbb{R}^{m}$ of the $j$-th sample 
% $\mathbf{X}^j$, its activation weights
is a function of input data by the gating $G(\cdot)$, i.e., 
$\bm{\beta}^j = G(\mathbf{X}^j)$.

Till now, MoE has been widely used in multi-task learning \cite{zheng2019self}, video captioning \cite{wang2019learning}, and multilingual neural machine translation (NMT) \cite{dai2022stablemoe} tasks. Taking NMT as an example, prior work \cite{dai2022stablemoe} shows that MoE can help learn a model from diverse sources (e.g., different languages), where they share the most common parameters but also specialize for individual % adaptive 
source via ``experts'' (sub-networks). 
% In this work, 
We combine the strengths of original and sparse MoE for our
\method.
% to design 
% \am layer.
% for AD.
\vspace{-0.05in}
% \subsubsection{3.4.1~~Repurposed Mixture-of-experts (MoE) Architecture for specialized and scalable AD.}
\subsubsection{3.3.2~Capitalizing Mixture-of-experts (MoE) Architecture for \prob.}

It is easy to see the \textit{\textbf{connection}} between MoE's applications (e.g., 
% multi-task learning and 
multilingual machine translation) and \prob---in both cases  
% they all try to 
MoE can help 
\textit{\textbf{learn from diverse sources}} (e.g., multiple sets of noisy labels in \prob) to capture the \textit{similarity 
via parameter sharing} while encouraging \textit{specialization via expert learning}. 

By recognizing this connection, 
% In this study, 
we (for the first time), introduce MoE architecture for weakly supervised AD with noisy labels (i.e., \prob), called \am. 
Similar to other works that use MoE, 
proposed \method 
% framework
keeps (does not alter) an (AD) algorithm's original layers  \textit{before} the output layer (e.g., dense layers in an MLP) 
to explore the agreement among noisy sources via shared parameters, while inserting an MoE layer before the output layer to learn from each noisy source with specialization (via the experts/sub-networks). In an ideal world, each expert is good at handling samples from different noisy sources and updated only with more accurate sets of noisy labels. As the toy example shown in Fig. \ref{fig:ours} to apply \method on an MLP for AD, we insert an \method layer between the dense layers and the output layer: where the \method layer contains four experts, and the gating activates only the top two for each input example. 

For the $j$-th sample, the MoE layer's output $O^j$ is shown in Eq. (\ref{eq:moe}) as a weighted sum of all activated experts' outputs, where $E_i(\cdot)$ denotes the $i$-th expert network, and $h^j$ is the output from the dense layer (as the input to \method). $\beta_i^j$ is the weight of the $i$-th 
% \textit{activated}
expert assigned by the gating function $G(\cdot)$, and we describe its calculation in Eq. (\ref{eq:new_weight}). Although Eq. (\ref{eq:moe}) enumerates all $m$ experts for aggregation, only the top-$k$ experts with non-negative weights $\beta_i^j>0$ are used.  

 \vspace{-0.075in}

\begin{equation}
    % \vspace{-0.025in}
    O^j = \sum_{i=1}^m \beta_i^j E_i(h^j)
    \label{eq:moe}
    % \vspace{-0.025in}
\end{equation}

\noindent \textbf{Improving \method with Noisy-Label Aware Expert Activation.} The original MoE calculates the activation weights with only the raw input feature $\mathbf{X}$ (see \S 3.3.1), which can be improved in \prob with the presence of noisy labels. 
% As an improvement, 
To such end,
we explicitly make the gating function $G(\cdot)$ aware of noisy labels $\mathcal{Y}_w$ while calculating the weights for (activating) experts.
Intuitively, $t$ sets of noisy labels can be expressed as $t$-element binary vectors (0 means normalcy and 1 means abnormality). However, it is hard for neural networks to directly learn binary inputs \citep{buckman2018thermometer}. Thus, we propose to learn a $\mathbb{R}^{t\times e}$ continuous embedding $Emb(\cdot)$ for noisy labels ($e$ is embedding dimension), and use both the raw input features $\mathbf{X}$ in $d$ dims and the average of the embedding in $e$ dims as the input of the gating.
% , resulting in noisy-label aware input in $\mathbb{R}^{(d+e)}$). 
As such, the expert weights for the $j$-th sample $\bm{\beta}^j$ can be calculated with Eq. (\ref{eq:new_weight}), and plugged back into Eq. (\ref{eq:moe}) for \method output.
\vspace{-0.05in}

\begin{equation}
    \bm{\beta}^j = G\left(\mathbf{X}^j, Emb(\mathcal{Y}_w^j)\right)
    \label{eq:new_weight}
\end{equation}

\noindent \textbf{Loss Function}. 
% It is crucial to remark 
% Note that 
\method is strictly an enhancement framework for \prob other than a new detection algorithm, and thus we do not need to design a new loss function but just enrich the original loss $\mathcal{L}_o$ of the underlying AD algorithm (e.g., cross-entropy for a simple MLP). While calculating the loss in each batch of data, we  \textit{randomly sample one set} of noisy labels treated as the (pseudo) clean labels or \textit{combine the loss of all $t$ sets} of noisy labels by treating them as (pseudo) clean labels. To encourage all experts to be evenly activated and not collapse on a few of them \cite{riquelme2021scaling}, we include an additional loss term on gating (i.e, $\mathcal{L}_g$). Putting these together, we show the loss for the $j$-th training sample in Eq. (\ref{eq:loss}), where the first term encourages equal activation of experts with $\alpha$ as the load balancing factor, and 
% \Circled{1} encourages all experts to be activated equally. The \Circled{2} sums over the all 
the second part 
% of the equation
% $\beta_i^j$ is the gating weight. $\mathcal{L}_{o}(\cdot)$ 
is the loss of the $j$-th sample, which depends on the MoE layer output $O^j$ in Eq. (\ref{eq:moe}) 
% (where $\sigma$ is the softmax) 
and the corresponding noisy labels $\mathcal{Y}_w^j$ (either one set or all in loss calculation).
% , and is loss of .
% \vspace{-0.025in}
\begin{equation}
    % \vspace{-0.025in}
    \mathcal{L}^j = 
    % \underbracket{
    \alpha \mathcal{L}_g
    % }_\textrm{\Circled{1}} 
    +
    % \sum_{l=1}^k
 % \underbracket{
 \mathcal{L}_{o}\left(O^j, \mathcal{Y}_w^j\right)
 % }
 % _\textrm{\Circled{3}}
    \label{eq:loss}
    % \vspace{-0.025in}
\end{equation}

\noindent \textbf{Remark on the Number of \method Layers and Sparsity.} Similar to the usage in NLP, one may use multiple \method layers in an AD model (e.g., every other layer), especially for the AD models with complex structures like transformers \cite{li2022self}. In this work, we only show inserting the \method layer before the output layer (see the illustration in Fig. \ref{fig:ours}) of MLP, while future work may explore more extensive use of \method layers in complex AD models. 
As AD models are way smaller \cite{pang2021deep}, 
we do not enforce gating sparsity as in NLP models \cite{MoE}. See ablation studies on the total number of experts and the number of activated experts in \S 4.4.3.
\vspace{-0.025in}
% In addition to perfectly fulfill \prob, 
% We summarize three 
% \noindent \textit{Combining the strength of original and sparse MoE for AD}:
\subsubsection{3.3.3~Advantages and Properties of \method\\}

\textbf{Implicit Mapping of Noisy Labels to Experts with Better Scalability}. 
% repurposing MoE for \prob 
The proposed \method (Fig. \ref{fig:ours}) is more scalable than existing noisy-label learning methods for classification, e.g., CrowdLayer (Fig. \ref{fig:crowdlayer}) and UnionNet (Fig. \ref{fig:unionnet}). As discussed in \S \ref{sec:related_b2}, these methods explicitly map each set of noisy labels to part of the network.
% , either to recover the labels or learn a transition matrix. 
With more sets of noisy labels, the network parameters designated for (recovering or mapping) noisy labels increase proportionately. In contrast, \method does not enforce explicit one-to-one mapping from noisy labels to experts. 
% Consequently, 
Thus,
its many-to-many mapping becomes more scalable with many noisy sources.

\noindent \textbf{Extension with Clean Labels}. Our design also allows for easy integration of clean labels. In many AD tasks, a small set of clean labels are feasible, where \method can easily use them. No update is needed for the network design, but simply treating the clean label as ``another set of noisy labels" with higher weights. See \S 4.4.4 for experiment results.

\vspace{-0.1in}

\section{Experiments}
\label{sec:exp}
We design experiments to answer the following questions:
\cben
    \item How do \method-enhanced methods compare to SOTA AD methods that learn from a single set of labels? (\S \ref{exp:leading_AD})
    \item How does \method compare to leading (multiple sets of) noisy label learning methods for classification? (\S \ref{exp:leading_nl})
    \item How does \method perform under different settings, e.g., varying num. of experts and clean label ratios? (\S \ref{exp:ablation})
\ceen

\subsection{Experiment Setting}
\label{subsec:exp_setting}

\begin{table}[!htb]
\centering
\footnotesize
	\footnotesize
	\vspace{-0.1in}
	\scalebox{0.75}{
	\begin{tabular}{l|lllll } % centered columns (4 columns)
	\toprule
	 \textbf{Data}                 & \textbf{\# Samples} & \textbf{\# Features} & \textbf{\# Anomaly} & \textbf{\% Anomaly} & \textbf{Category} \\
	\midrule
ag news                           & 10000    & 768       & 500       & 5.00              & NLP\\
aloi                    & 49534   & 27       & 1508      & 3.04                &     Image     \\
mnist                                & 7603    & 100      & 700       & 9.21                & Image         \\

spambase            & 4207    & 57       & 1679      & 39.91               & Doc         \\
svhn                           & 5208    & 512       & 260       & 5.00               & Image \\

Imdb                           & 10000    & 768       & 500       & 5.00               & NLP\\
Yelp                          & 10000    & 768       & 500       & 5.00               & NLP\\
\midrule
security$^*$ & 5525 & 21 & 378 & 6.84 & Security\\

    \bottomrule
	\end{tabular}
	}
 \vspace{-0.05in}
	\caption{Data description of the eight datasets used in this study: the top seven datasets are adapted from AD repo., e.g., DAMI \cite{Campos2016} and ADBench \cite{han2022adbench}, and security$^*$ is a proprietary enterprise-security dataset.} % title of Table
	\label{table:datasets} % is used to refer this table in the text
	\vspace{-0.1in}
\end{table}

\textbf{Benchmark Datasets}. As shown in Table \ref{table:datasets}, we evaluate \method on seven public datasets adapted from AD repositories \cite{Campos2016,han2022adbench} and 
a proprietary enterprise-security dataset (with $t=3$ sets of noisy labels). 
Note that these public datasets do not have existing noisy labels for AD, so we simulate $t=4$ sets of noisy labels per dataset via two methods:
\cben
\item \textbf{Label Flipping} \citep{zheng2021meta} generates noisy labels by uniformly swapping anomaly and normal classes at a designated noise rate.
\item \textbf{Inaccurate Output} uses varying percentages of ground truth labels to train $t$ diverse classifiers, and considers their (inaccurate) predictions as noisy labels. With more ground truth labels to train a classifier, its prediction (e.g., noisy labels) will be more accurate---we control the noise levels by the availability of ground truth labels.
\ceen

In this work, we use the noisy labels simulated by \textbf{Inaccurate Output} since that is more realistic and closer to real-world applications (e.g., noise is not random), while we also release the datasets by \textbf{Label Flipping} for broader usage.
% \textcolor{blue}{Explain why we use the later}
We provide a detailed dataset description in Appx. \ref{appx:datasets}.

\noindent \textbf{Two Groups of Baselines} are described below: 
% We introduce \textit{two groups of baselines}.
% in this study. 
\cben
\item 
% First, we compare \method with 
\textit{Leading AD methods that can only handle a single set of labels} to show \method's benefit of leveraging multiple sets of noisy labels. These include 
\textit{SOTA AD methods}: \textbf{(1) XGBOD} \citep{zhao2018xgbod}  \textbf{(2) PreNet} \citep{pang2019deep} \textbf{(3) DevNet} \citep{pang2019devnet}
\textbf{(4) DeepSAD} \cite{ruff2019deep}, and 
% \item 
\textit{popular classification methods}: \textbf{(5) MLP} \textbf{(6) XGBoost} \citep{chen2016xgboost} \textbf{(7) LightGBM} \citep{ke2017lightgbm}. We provide detailed descriptions in Appx. \S \ref{appx:semiad}.
\item 
% Additionally, we compare \method against 
\textit{Leading methods (for classification) that handle multiple sets of noisy labels} (we adapt them for \prob; see \S \ref{sec:related_b2} and Table \ref{table:baselines} for details): 
\textbf{(1) SingleNoisy}
\textbf{(2) LabelVote} \textbf{(3) HyperEnsemble} \cite{wenzel2020hyperparameter} that averages $t$ models' scores (referred as \textbf{HP\_A}) or takes their max (referred as \textbf{HP\_M})  
\textbf{(4) CrowdLayer} \cite{rodrigues2018deep} 
% (4) \textbf{DoctorNet (DN)} \cite{guan2018said}
% (5) \textbf{WDN} \cite{guan2018said} 
and latest 
\textbf{(5) UnionNet} \cite{wei2022deep}.
% \ceit
\ceen

\begin{figure} [!b]
    \centering
    \vspace{-0.1in}
	\subfloat[\texttt{agnews}] {%
			\includegraphics[clip,width=0.48\columnwidth]{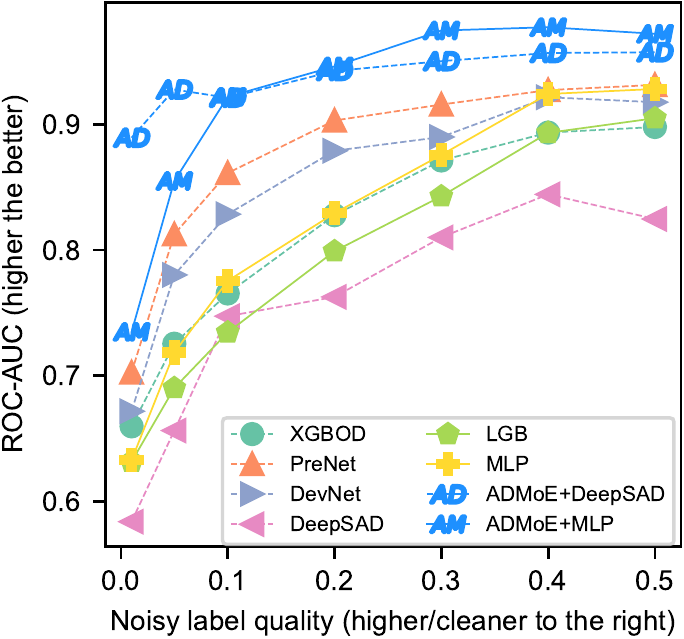}%
		\label{fig:agnews}
	}
	\subfloat[\texttt{aloi}] {%
			\includegraphics[clip,width=0.48\columnwidth]{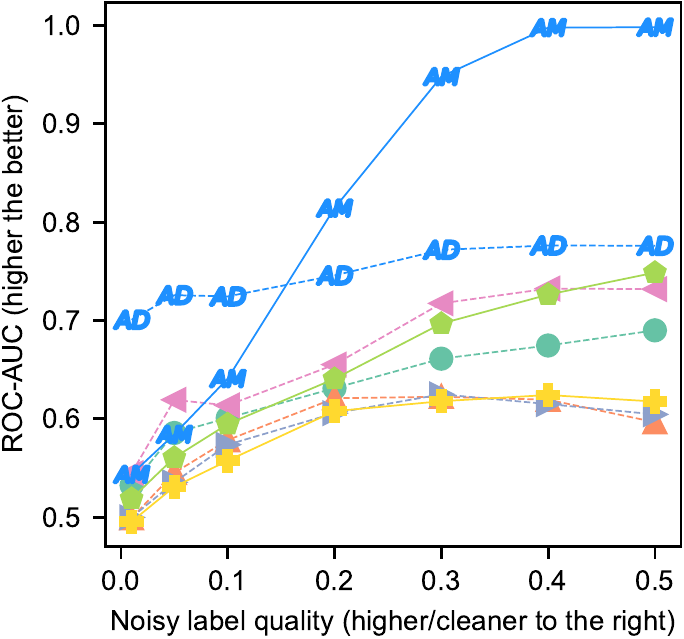}%
		\label{fig:aloi}
	}
	
% 	\hspace{0.02in}
	\subfloat[\texttt{imdb}]{%
		\includegraphics[clip,width=0.48\columnwidth]{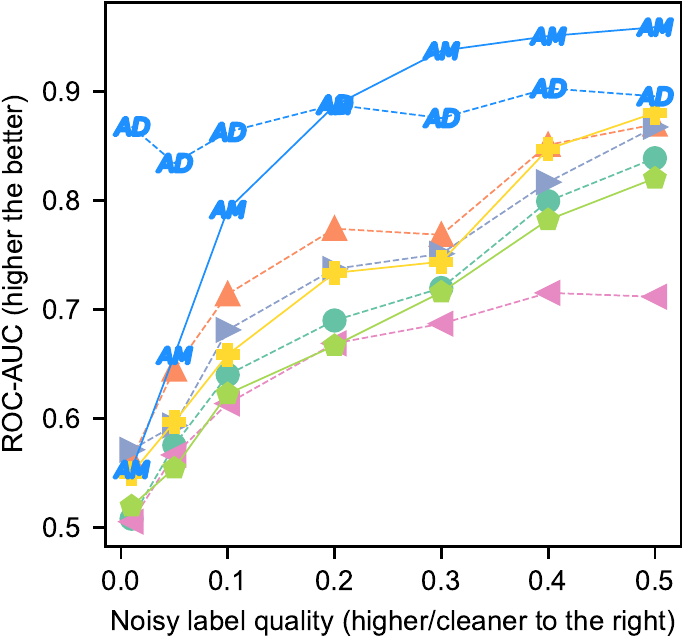}%
	\label{fig:imdb}
	}
	\subfloat[\texttt{mnist}] {%
		\includegraphics[clip,width=0.48\columnwidth]{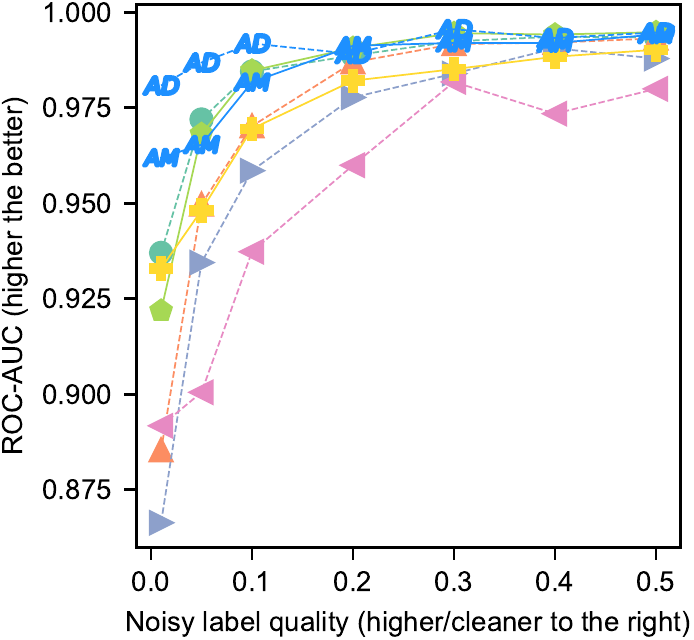}%
		\label{fig:security}
	}
	
% 	\hspace{0.02in}
	\subfloat[\texttt{spamspace}] {%
		\includegraphics[clip,width=0.48\columnwidth]{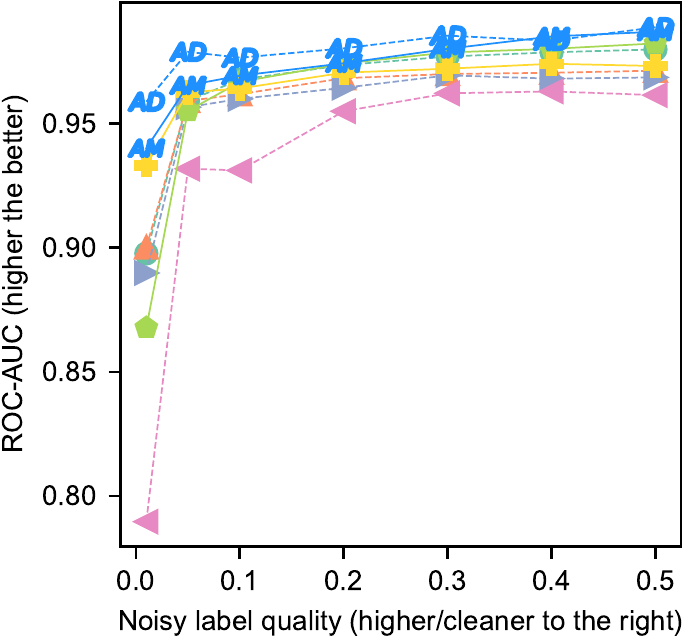}%
		\label{fig:spamspace}
	}
% 	\hspace{0.02in}
	\subfloat[\texttt{svhn}] {%
		\includegraphics[clip,width=0.48\columnwidth]{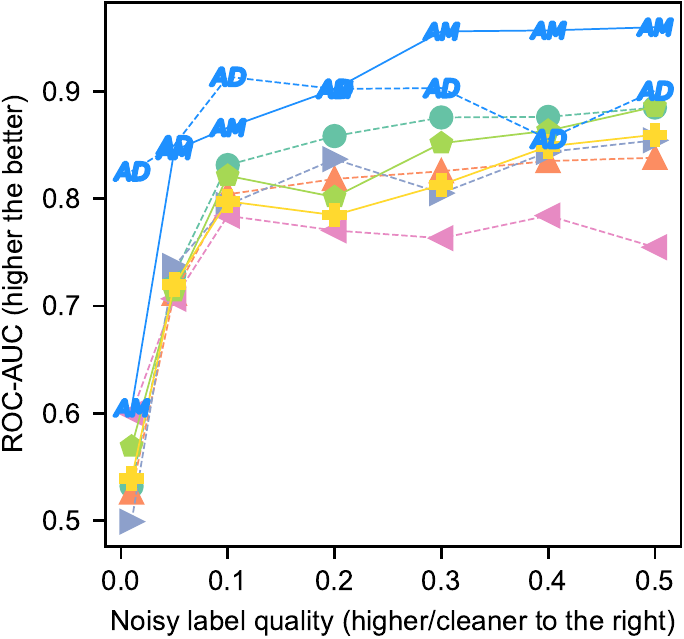}%
		\label{fig:svhn}
	}

	\vspace{-0.1in}
	\caption{
	ROC-AUC comparison at different noisy label quality of \method enhanced DeepSAD and MLP (denoted as \textcolor{dodgerblue}{\textbf{\textit{AD}}} and \textcolor{dodgerblue}{\textbf{\textit{AM}}}) with leading AD methods that can only leverage one set of noisy labels. We already show \texttt{Yelp}'s result in Fig. \ref{fig:mlp_demo}. Notably, \method enabled \textcolor{dodgerblue}{\textbf{\textit{AD}}} and \textcolor{dodgerblue}{\textbf{\textit{AM}}} has significant performance improvement especially when the label quality is very low (to the left of the x-axis). 
	}
	\vspace{-0.2in}
	\label{fig:ad_baseline_comp}
\end{figure}
\begin{table*}[!htp]
\centering
	\footnotesize
	\vspace{-0.1in}
	\centering
\scalebox{0.9}{
\begin{tabular}{l|cccc|ccc|ccc}
\toprule
% &&&&& \multicolumn{3}{c}{MLP}&&& \\
\textbf{Dataset}   & \textbf{XGBOD} & \textbf{PreNet} & \textbf{DevNet} & \textbf{LGB} & \textbf{MLP} & \makecell{\textbf{\method-}\\ \textbf{MLP}} & \textbf{$\Delta$ Perf.} & \textbf{DeepSAD} & \makecell{\textbf{\method-}\\ \textbf{DeepSAD}} & \textbf{$\Delta$ Perf.} \\
\midrule
\textbf{agnews}    & 0.8277         & 0.903           & 0.8789          & 0.7991       & 0.8293       & \textbf{0.946}  & +14.07\%             & 0.7627           & 0.9423          & +23.55\%            \\
\textbf{aloi}      & 0.6312         & 0.6209          & 0.6057          & 0.6406       & 0.6078       & \textbf{0.8142} & +33.96\%             & 0.6552           & 0.7458          & +13.83\%            \\
\textbf{imdb}      & 0.6898         & 0.7741          & 0.7372          & 0.6667       & 0.7335       & 0.8872          & +20.95\%             & 0.6688           & \textbf{0.8881} & +32.79\%            \\
\textbf{mnist}     & 0.9887         & 0.9869          & 0.9777          & 0.9907       & 0.9821       & \textbf{0.9913} & +0.94\%              & 0.96             & 0.9891          & +3.03\%             \\
\textbf{spamspace} & 0.9736         & 0.9683          & 0.9645          & 0.9743       & 0.9706       & 0.9744          & +0.39\%              & 0.955            & \textbf{0.9804} & +2.66\%             \\
\textbf{svhn}      & 0.8585         & 0.8185          & 0.837           & 0.8016       & 0.7847       & 0.9022          & +14.97\%             & 0.7705           & \textbf{0.9025} & +17.13\%            \\
\textbf{yelp}      & 0.7778         & 0.8699          & 0.8689          & 0.7348       & 0.7955       & \textbf{0.9535} & +19.86\%             & 0.7781           & 0.9355          & +20.23\%            \\
\midrule
\textbf{security$^{*}$}  & 0.7479         & 0.7415          & 0.7498          & 0.7335       & 0.7363       & 0.7767          & +5.49\%              & 0.7928           & \textbf{0.8108} & +2.27\%              \\
\midrule
\midrule
\textbf{Average} & 0.8119 & 0.8353 & 0.8274 & 0.7926 & 0.8049 & \textbf{0.9056} & +13.83\% & 0.7929 & 0.8993 & +14.44\% \\  
\bottomrule
\end{tabular}
}
\vspace{-0.05in}
\caption{Performance comparison between \method-enhanced AD methods and leading AD methods (that can only use one set of labels) at noisy level 0.2. The best performance is highlighted in bold per dataset (row). \method-based methods (\method-MLP and \method-DeepSAD denote \method-enhanced MLP and DeepSAD) outperform all baselines. \method brings on average 13.83\% and up to 33.96\% improvement over the original MLP, and on average 14.44\% and up to 32.79\% improvement over DeepSAD. Note that all the neural-network models use the equivalent numbers of parameters and training FLOPs.} % title of Table

	\label{table:ad_results} % is used to refer this table in the text
	\vspace{-0.2in}
\end{table*}

\noindent \textbf{Backbone AD Algorithms, Model Capacity, and Hyperparameters}. We show the generality of \method to enhance (\textit{i}) simple MLP and (\textit{ii}) SOTA DeepSAD \cite{ruff2019deep}. To ensure a fair comparison, we ensure all methods have the equivalent number of trainable parameters and FLOPs. 
See Appx. \ref{appx:hp} and code for additional settings (e.g., hyperparameters) in this study.
All experiments are run on an Intel i7-9700 @3.00 GH, 64GB RAM, 
8-core workstation with an NVIDIA Tesla V100 GPU.  

\noindent \textbf{Evaluation}. For methods with built-in randomness, we run four independent trials and take the average, with a fixed dataset split (70\% train, 25\% for test, 5\% for validation). Following AD research tradition \cite{aggarwal2017outlier,zhao2019lscp,lai2021tods,han2022adbench}, we report ROC-AUC as the primary metric, while also showing additional results of average precision (AP) in Appx. \ref{appx:ap}.
\vspace{-0.1in}

\subsection{Comparison Between \method
% Enhanced 
Methods and Leading AD Methods (Q1)}
\label{exp:leading_AD}

 Fig. \ref{fig:mlp_demo} and Fig. \ref{fig:ad_baseline_comp} show that \textbf{\method enables MLP and DeepSAD to use multiple sets of noisy labels, which outperform leading AD methods that can use only one set of labels}, at varying noisy label qualities (x-axis). We further use Table \ref{table:ad_results} to compare them at noisy label quality 0.2 to understand the specific gain of \method using multiple sets of noisy labels. The third block of the table shows that \method brings on average 13.83\%, and up to 33.96\% (\texttt{aloi}; 3-rd row) improvements to a simple MLP. Additionally, the fourth block of the table further demonstrates that \method enhances DeepSAD for using multiple noisy labels, with on average 14.44\%, and up to 32.79\% (\texttt{imdb}; 4-th row) gains. These results demonstrate the benefit of using \method to enable AD algorithms to learn from multiple noisy sources. 

\noindent \textbf{\method shows larger improvement when labels are more noisy} (to the left of x-axis of Fig. \ref{fig:ad_baseline_comp}). We observe that \method-DeepSAD brings up to 60\% of ROC-AUC improvement over the best-performing AD algorithm at noisy label quality 0.01 (see Fig. \ref{fig:imdb} for \texttt{imdb}). This observation is meaningful as we mostly need an approach to improve detection quality when the labels are extremely noisy, where \method yields more improvement. Fig. \ref{fig:ad_baseline_comp} also demonstrates when the labels are less noisy (to the right of the x-axis), the performance gaps among methods are smaller since the diversity among noisy sources is also reduced---using any set of noisy labels is sufficient to train a good AD model. Also, note that \method-DeepSAD shows better performance than \method-MLP with more noisy labels. This results from the additional robustness of the underlying semi-supervised method DeepSAD with access to unlabeled data.
% \vspace{-0.2in}

\subsection{Comparison Between \method and Noisy Label Learning Methods (Q2)}
\label{exp:leading_nl}

\begin{figure} [!t]
    \centering
% 	\hspace{0.02in}
	\subfloat[\texttt{imdb}]{%
		\includegraphics[clip,width=0.48\columnwidth]{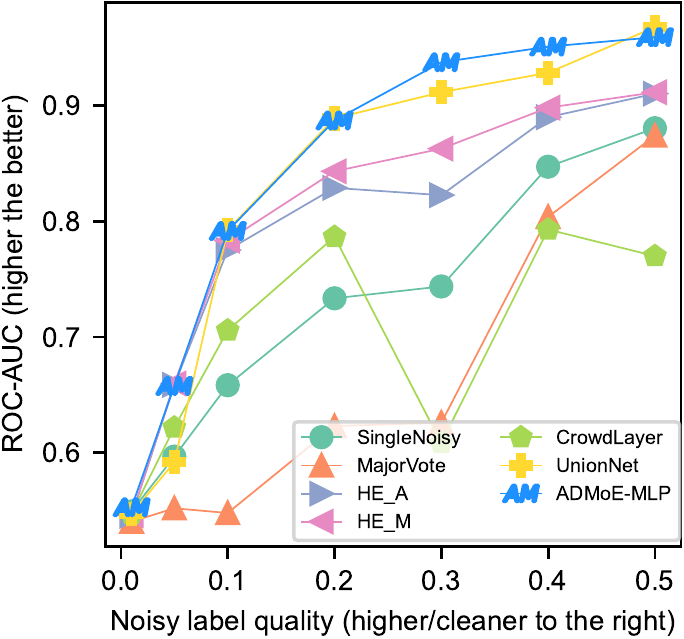}%
	\label{fig:imdb_nl}
	}
% 	\hspace{0.02in}
	\subfloat[\texttt{svhn}] {%
		\includegraphics[clip,width=0.48\columnwidth]{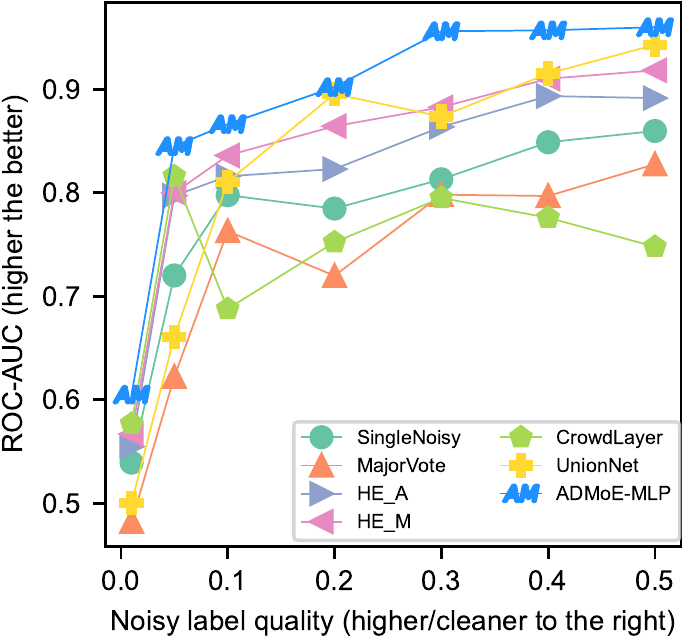}%
		\label{fig:svhn_nl}
	}

	\vspace{-0.1in}
	\caption{
	ROC-AUC comparison at different noisy label quality of \method enhanced MLP (\textcolor{dodgerblue}{\textbf{\textit{AM}}}) with leading multi-set noisy-label learning methods. We already show \texttt{Yelp}'s result in Fig. \ref{fig:perf_highlight}. \method outperforms most baselines.
	}
	\vspace{-0.1in}
	\label{fig:nl_baseline_comp}
\end{figure}

\begin{table}[!htp]
\centering

	\footnotesize
	% \vspace{-0.1in}
\scalebox{0.74}{
\begin{tabular}{l|llllll|l}
\toprule
\textbf{Dataset}   & \makecell{\textbf{Single}\\\textbf{Noisy}} & \makecell{\textbf{Major}\\\textbf{Vote}} & \textbf{HE\_A}   & \textbf{HE\_M} & \makecell{\textbf{Crowd}\\\textbf{Layer}} & \makecell{\textbf{Union}\\\textbf{Net}} & \textbf{\method}   \\
\midrule
\textbf{agnews}    & 0.7185               & 0.5729             & 0.7824          & 0.8101        & 0.8543              & 0.7412            & \textbf{0.8549} \\
\textbf{aloi}      & 0.531                & 0.479              & 0.5195          & 0.5579        & 0.5495              & 0.5773            & \textbf{0.5842} \\
\textbf{imdb}      & 0.5967               & 0.5521             & 0.6587 & \textbf{0.6599}        & 0.6219              & 0.5918            & 0.6577          \\
\textbf{mnist}     & 0.9482               & 0.9467             & 0.9598          & 0.9563        & 0.9644              & 0.6767            & \textbf{0.9655} \\
\textbf{spamspace} & 0.9626               & \textbf{0.9655}    & 0.9584          & 0.9586        & 0.7654              & 0.9567            & \textbf{0.9655} \\
\textbf{svhn}      & 0.7201               & 0.6222             & 0.7971          & 0.7998        & 0.8161              & 0.6608            & \textbf{0.8451} \\
\textbf{yelp}      & 0.6777               & 0.5708             & 0.7298          & 0.7358        & 0.7418              & 0.7133            & \textbf{0.7533} \\
\midrule
\textbf{security$^{*}$}  & 0.7363               & 0.7653             & 0.7526          & 0.7484        & 0.7374              & 0.7435            & \textbf{0.7767} \\
\midrule
\midrule
\textbf{Average} &0.7363 &	0.6843	&0.7722	&0.7783&	0.7563	&0.7014&	\textbf{0.8003} \\
\bottomrule
\end{tabular}
}
\vspace{-0.1in}
	\caption{Performance comparison between \method and leading noisy-label learning methods (in Table \ref{table:baselines}) on 
 % enhancing 
 an MLP at noisy label quality 0.05. The best performance is highlighted in bold per dataset (row). \method mostly outperforms baselines, with on avg. 9.4\% and up to 19\% improvement over SingleNoisy which trains 
 % an MLP 
 w/ a set of noisy labels.} % title of Table

	\label{table:nl_results} % is used to refer this table in the text
\vspace{-0.15in}
\end{table}

\textbf{\method also shows an edge over SOTA \textit{classification} methods that learn from multiple sets of noisy labels\footnote{We adapt these classification methods (Table \ref{table:baselines}) for \prob.}}, as shown in Fig. \ref{fig:perf_highlight} and \ref{fig:nl_baseline_comp}. We also analyze the comparison at noisy label quality 0.05 in Table \ref{table:nl_results}. where \method ranks the best in 7 out of 8 datasets. In addition to better detection accuracy, \method only builds a single model, and is thus faster than \textbf{HE\_E} and \textbf{HE\_M} which requires building $t$ independent models to aggregate predictions. We credit \method's performance over SOTA algorithms including CrowdLayer and UnionNet to its implicit mapping of noisy labels to experts
% for better specialization 
(\S 3.3.3).

\subsection{Ablation Studies and Other Analysis (Q3)}
\label{exp:ablation}

\subsubsection{4.4.1. Case Study: How does \method Help?}
\label{subsec:case}
% In NLP literature, 
Although MoE is effective in various NLP tasks, it is often challenging to contextualize how each expert responds to specific groups of samples due to the complexity of NLP models \cite{MoE,zheng2019self,zuo2021taming}. Given that AD models are much smaller and we use only one MoE layer before the output layer, we present an interesting case study (MLP with 4 experts where top 1 gets activated) in Table \ref{table:moe_case}. We find each expert achieves the highest ROC-AUC on the subsamples where it gets activated by gating (see the diagonal), and they are significantly better than training 
an individual model with the same architecture (see the last column). Thus, each expert does develop a specialization in the subsamples they are ``responsible'' for. 
% See ablation studies on using MoE in \S 4.4.1.

\begin{table}[!t]
\centering
 % title of Table
	\footnotesize
	% \vspace{-0.1in}
\scalebox{0.73}{
\begin{tabular}{c|llll|l}
\toprule
\textbf{Activated Expert}  &  \textbf{Expert 1} & \textbf{Expert 2} & \textbf{Expert 3} & \textbf{Expert 4} & \textbf{Comp.}\\
\midrule
\textbf{Subsamples for Expert 1} & \textbf{0.8554}                            & 0.8488                         & 0.8458                             & 0.8440                            & 0.7741                            \\
\textbf{Subsamples for Expert 2} & 0.8995                             & \textbf{0.9043 }                            & 0.8913                             & 0.8943                             & 0.7976                             \\
\textbf{Subsamples for Expert 3} & 0.8633                             & 0.8643                             & \textbf{0.8729}                             & 0.8665                             & 0.8049                             \\
\textbf{Subsamples for Expert 4} & 0.7903                             & 0.7888                             & 0.7775                             & \textbf{0.8066}                             & 0.7134 \\           
\bottomrule
\end{tabular}
}
\vspace{-0.05in}
	\caption{Perf. breakdown of each expert and a comparison model (w/ the same capacity as each expert but trained independently; last col.) on subsamples activated for each expert by MoE on \texttt{Yelp}. We highlight the best model per row in \textbf{bold}. The specialized expert performs the best in their assigned subsamples by gating, i.e., the diagonal is all in bold.}
	\label{table:moe_case} % is used to refer this table in the text
	\vspace{-0.2in}
\end{table}

\subsubsection{4.4.2~~Ablation on Using MoE and Noisy Labels as Inputs in \method.}
\begin{figure} [!ht]
    \centering
	\subfloat[\texttt{agnews}] {%
			\includegraphics[clip,width=0.48\columnwidth]{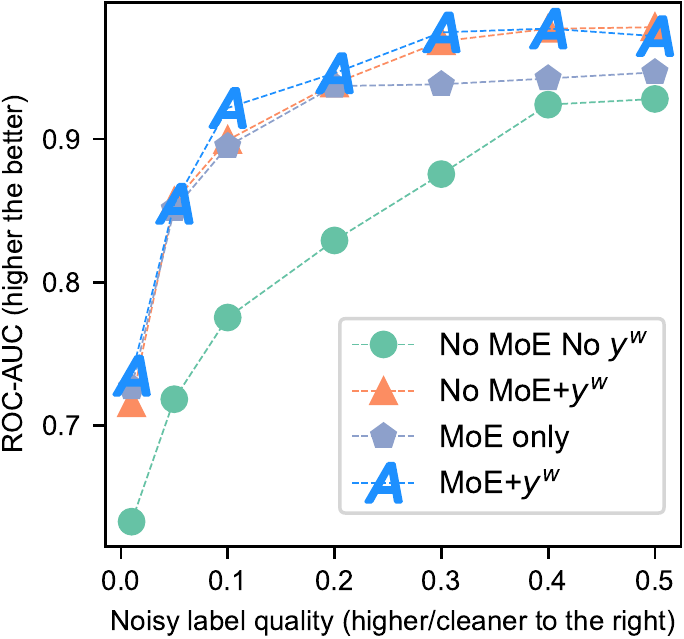}%
		\label{fig:ablation_agnews}
	}
	\subfloat[\texttt{imdb}] {%
			\includegraphics[clip,width=0.48\columnwidth]{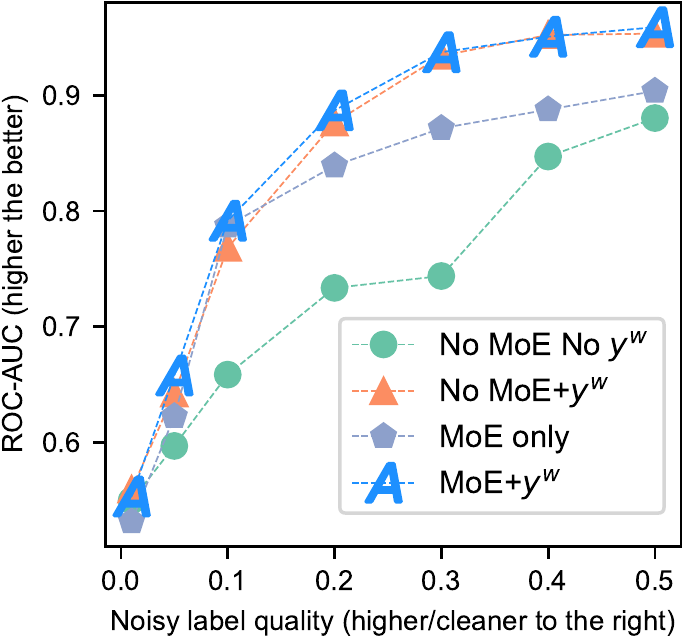}%
		\label{fig:ablation_imdb}
	}
	\vspace{-0.1in}
	\caption{
	Ablation studies on (\textit{i}) the use of \method layer and (\textit{ii}) the noisy labels $\mathbf{y}_w$ as input. \method (\textcolor{dodgerblue}{\textbf{\textit{A}}}) using both techniques shows the best results at (nearly) all settings.
 % noisy label qualities. 
	}
	\label{fig:ablation_moe_input}
	\vspace{-0.2in}
\end{figure}

% In addition to the case study, 
Additionally,
we analyze the effect of using (\textit{i}) \method layer and (\textit{ii}) noisy labels as input features in Fig. \ref{fig:ablation_moe_input} and Appx. Fig. \ref{fig:appx_ablation_moe_input}. First, \method performs the best while using these two techniques jointly in most cases, and significantly better than not using them (\mycircle{SeaGreen}). Second, \method helps the most when labels are noisier (to the left of the x-axis) with an avg. of 2\% improvement over only using noisy labels as input (\mytriangle{orange}). As expected, its impact is reduced with less noisy labels (i.e., closer to the ground truth): 
% when noisy labels are all high quality 
in that case, noisy labels are more similar to each other and specialization with \method is therefore limited.
\vspace{-0.05in}

\subsubsection{4.4.3~~Effect of Number of Experts ($m$) and top-$k$ Gating.}

\begin{figure} [!htp]
    \centering
	\subfloat[\texttt{agnews}] {%
			\includegraphics[clip,width=0.49\columnwidth]{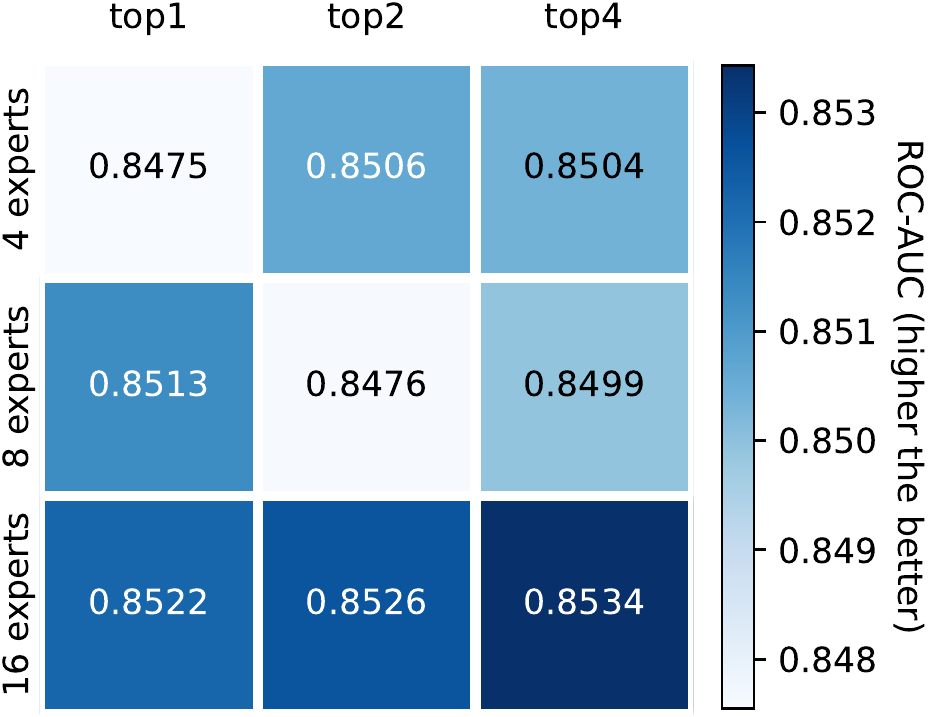}%
		\label{fig:expert_agnews}
	}
% 	\hspace{0.02in}
	\subfloat[\texttt{imdb}]{%
		\includegraphics[clip,width=0.49\columnwidth]{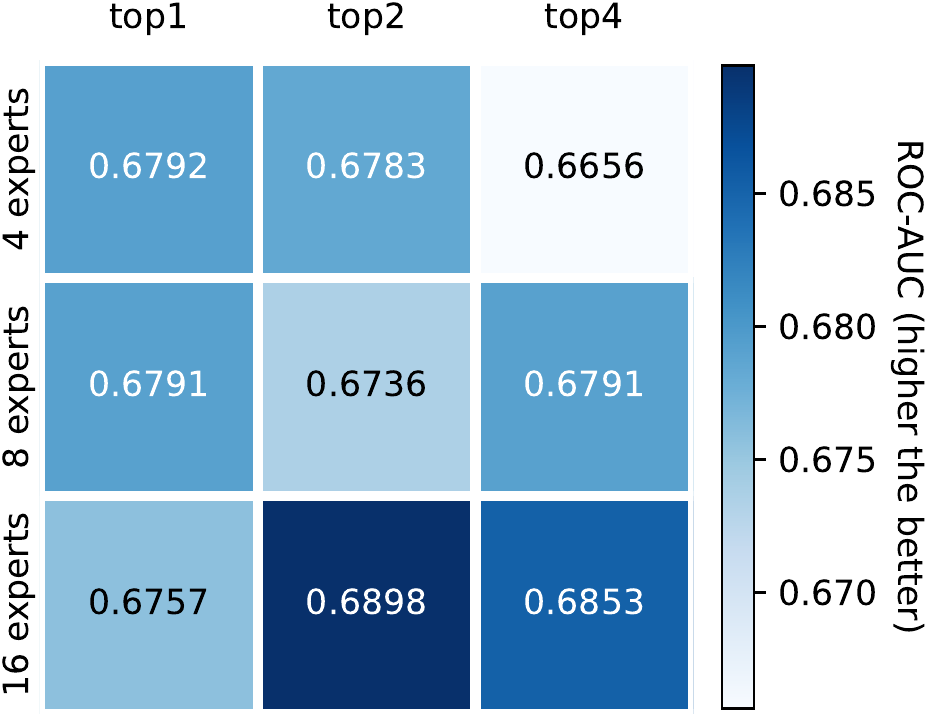}%
	\label{fig:expert_imdb}
	}
 
	\subfloat[\texttt{mnist}] {%
			\includegraphics[clip,width=0.49\columnwidth]{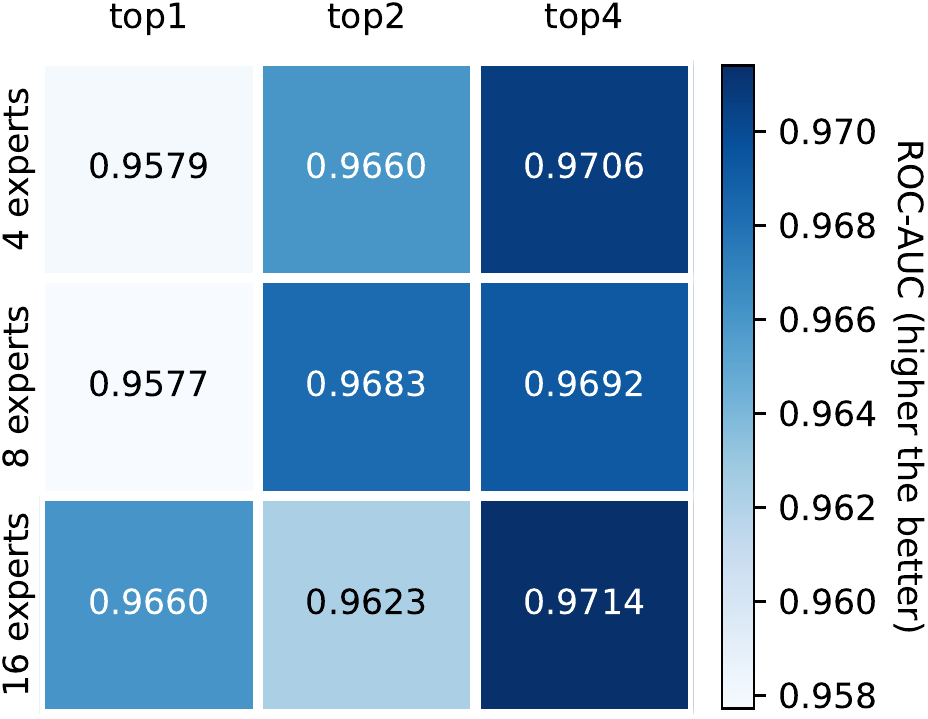}%
		\label{fig:expert_mnist}
	}
% 	\hspace{0.02in}
	\subfloat[\texttt{spamspace}] {%
		\includegraphics[clip,width=0.49\columnwidth]{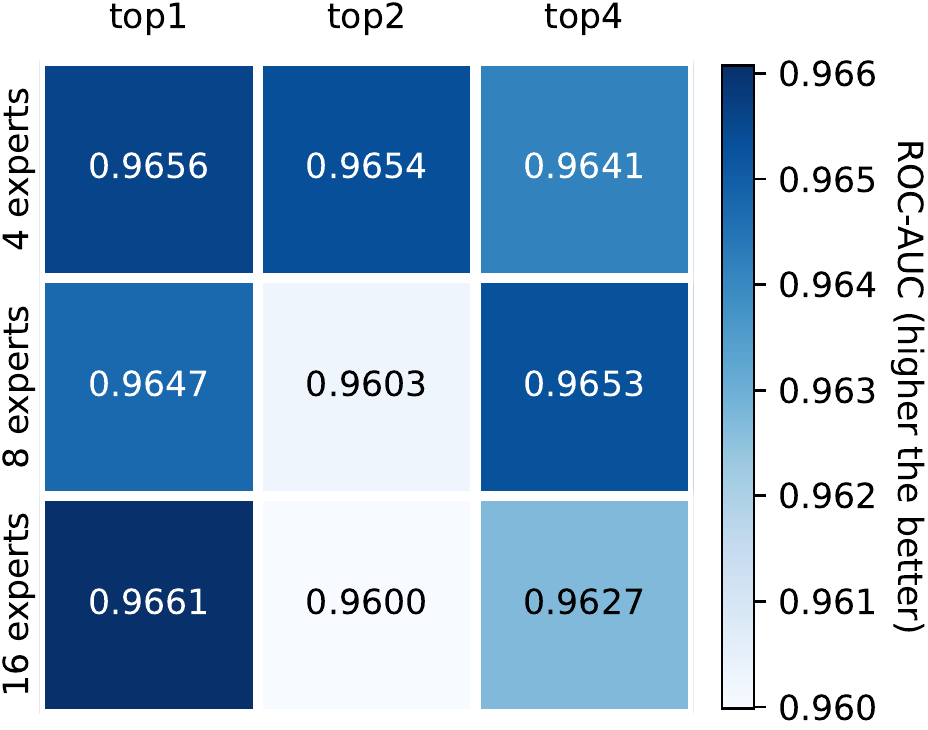}%
		\label{fig:expert_spamspace}
	}
	\vspace{-0.1in}
	\caption{
	Ablation studies on key hyperparameters in \method: (x-axis) the number of experts and (y-axis) top-$k$ experts to activate. We show the results at noisy level 0.05, and find 
 the best setting is data-dependent.
 See Appx. Fig. \ref{fig:ablation_num_experts_appx}.
 % \method's performance is insensitive to these HPs.
	}
	\vspace{-0.2in}
	\label{fig:ablation_num_experts}
\end{figure}

We vary the number of activated (x-axis) and total (y-axis) experts in \method, and compare their detection accuracy in Fig. \ref{fig:ablation_num_experts} and Appx. Fig. \ref{fig:ablation_num_experts_appx}. 
The results suggest that the best choice is data-dependent, and we use $m=4$ and $k=2$ for all the datasets in the experiments.
% To sum up, \method is insensitive to MoE's setting, and the best setting is data-dependent.

\subsubsection{4.4.4~~Performance on Varying Number of Clean Labels.} 
\begin{figure} [!htp]
    \centering
	\subfloat[\texttt{spamspace}] {%
		\includegraphics[clip,width=0.49\columnwidth]{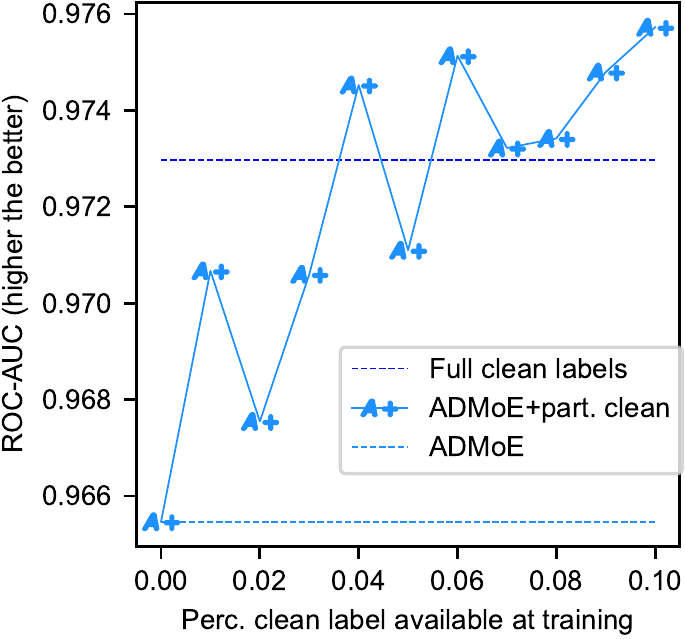}%
		\label{fig:clean_spamspace}
	}
	\subfloat[\texttt{svhn}] {%
		\includegraphics[clip,width=0.49\columnwidth]{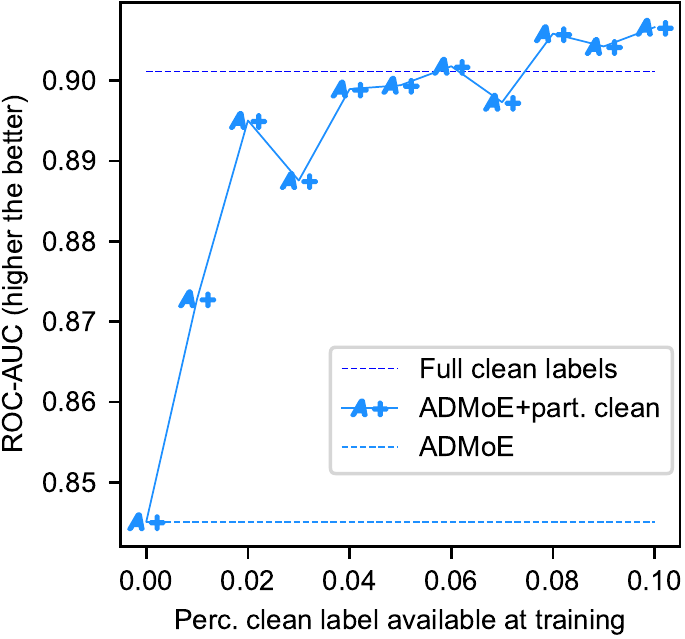}%
		\label{fig:clean_svhn}
	}
	\vspace{-0.1in}
	\caption{
	Analysis of integrating varying percentages (from 1\% to 10\%) of \textit{additional} clean labels in \method. The results show that \method efficiently leverages the clean labels with increasing performance. See more in Appx. Fig. \ref{fig:ablation_clean_appx}.
	}
	\vspace{-0.15in}
	\label{fig:ablation_clean}
\end{figure}

As introduced in \S 3.3.3, one merit of \method is the easy integration of 
% (a small set of) 
clean labels when available. Fig. \ref{fig:ablation_clean} and Appx. Fig. \ref{fig:ablation_clean_appx} show that \method can leverage the available clean labels to achieve higher performance. Specifically, \method with only 8\% clean labels can achieve similar or better results than that of using all available clean labels on \texttt{spamspace} and \texttt{svhn}.
\vspace{-0.05in}

\vspace{-0.1in}

\section{Conclusions}
\label{sec:conclusion}
We propose \method, a model-agnostic learning framework, to enable anomaly detection algorithms to learn from multiple sets of noisy labels. Leveraging Mixture-of-experts (MoE) architecture from the NLP domain, \method is scalable 
in building specialization based on the diversity among noisy sources. Extensive experiments show that \method outperforms a wide range of leading baselines, bringing on average 14\% improvement over not using it. Future work can explore its usage with complex detection algorithms.
\vspace{-0.1in}

\clearpage
\newpage

% \bibliographystyle{aaai23}
\bibliography{ref}

\clearpage
\newpage

\appendix 
\section*{Supplementary Material of \method}
\textit{Details on algorithm design, experiment setting, and additional results.}
\setcounter{table}{0}
\setcounter{figure}{0}

\renewcommand{\thetable}{\Alph{section}\arabic{table}}
\renewcommand{\thefigure}{\Alph{section}\arabic{figure}}
\renewcommand{\thealgorithm}{\Alph{section}\arabic{algorithm}}

\section{Additional Results on Why Does AD Benefit from Multiple Sets of Noisy Labels}
\label{appx:prob_examples}

\begin{figure} [h]
    \centering
	\subfloat[Result analysis on PreNet]{%
		\includegraphics[clip,width=0.48\columnwidth]{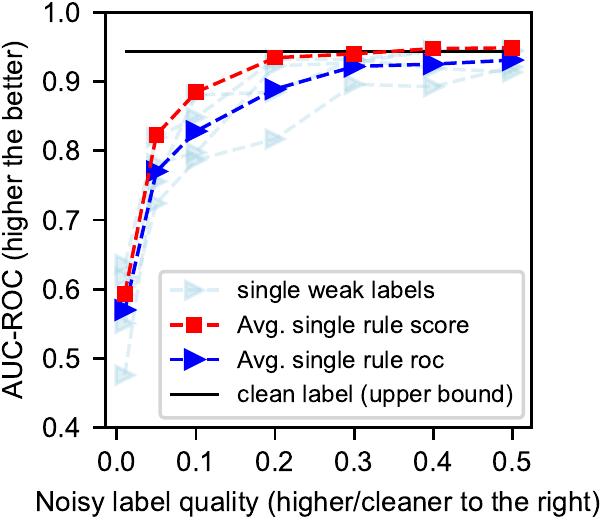}%
	\label{fig:perf_ensemble_prenet}
	}
	\hspace{0.02in}
% 	\vspace{0.1in}
	\subfloat[Result analysis on XGBOD] {%
		\includegraphics[clip,width=0.48\columnwidth]{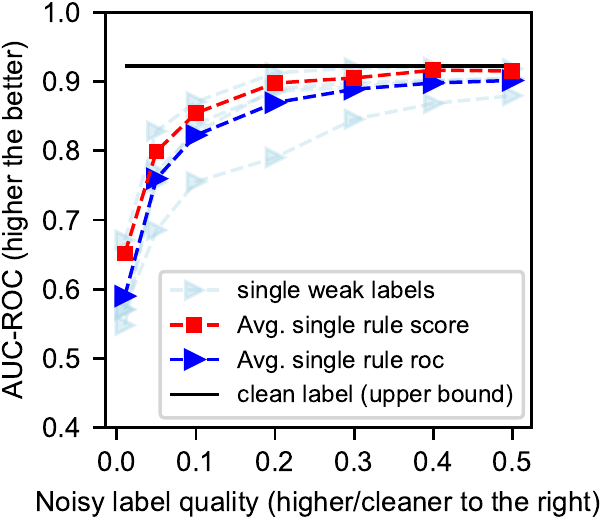}%
		\label{fig:perf_ensemble_xgbod}
	}
	\vspace{-0.1in}
	\caption{Benefit of leveraging multiple noisy sources on \texttt{Yelp}: even simply averaging individual models' outputs (\mysquare{red}) is better than training each weak source independently (\mytriangle{blue})}
	\vspace{-0.25in}
	\label{fig:perf_ensemble}
\end{figure}

\section{Additional Experiment Setting and Results}

\subsection{Dataset Description}
\label{appx:datasets}

As discussed in \S \ref{subsec:exp_setting}, we use the results based on the \textbf{Classification noise} since that is more realistic and closer to real-world applications (e.g., noise is not at random), while we also release the datasets by \textbf{Label flipping} for broader usages. See our repo to access both versions of datasets.

\textbf{Process of Inaccurate Output.} We use varying percentages of ground truth labels to train $t$ diverse classifiers (called noisy label generators), and consider their (inaccurate) predictions as noisy labels. Naturally, with more available ground truth labels to train a classifier, its prediction (e.g., noisy labels) will be more accurate---we therefore control the noise levels by the availability of ground truth labels.

Following this approach, we generate the noisy labels for the benchmark datasets at varying percentages of ground truth labels (namely, $\{0.01, 0.05, 0.1, 0.2, 0.3, 0.4, 0.5\}$) to train classifiers (a simple feed-forward MLP \cite{rosenblatt1958perceptron}, decision tree \cite{breiman2017classification}, and ensemble methods Random Forest \cite{breiman2001random} and LightGBM \cite{ke2017lightgbm}). Please see our code for more details.

Table \ref{table:label_roc} summarizes avg. ROC-AUC of the generated noisy labels while using varying percentages of ground truth labels to train noisy label generators. Of course, with more ground truth labels, the generated noisy labels are also more accurate. Note that \texttt{security$^*$} comes with actual noisy labels, and thus ROC-AUC does not vary.

\begin{table}[!ht]
\centering
\footnotesize
	\footnotesize
	\vspace{-0.1in}
	\scalebox{0.7}{
\begin{tabular}{c|lllllll|l}
\toprule
\makecell{\textbf{Clean}\\\textbf{Perc.}} & \textbf{agnews} & \textbf{aloi} & \textbf{imdb} & \textbf{mnist} & \makecell{\textbf{spam}\\\textbf{space}} & \textbf{svhn} & \textbf{yelp} & \textbf{security$^*$} \\
\midrule
\textbf{0.01}         & 0.5384          & 0.5154        & 0.5085        & 0.7260         & 0.7862             & 0.5164        & 0.5192        & 0.6862            \\
\textbf{0.05}                             & 0.5663          & 0.5330        & 0.5304        & 0.7887         & 0.8929             & 0.5773        & 0.5455        & 0.6862            \\
\textbf{0.1}                              & 0.6210          & 0.5540        & 0.5628        & 0.8592         & 0.9047             & 0.6132        & 0.5754        & 0.6862            \\
\textbf{0.2}                              & 0.6641          & 0.5957        & 0.6025        & 0.9044         & 0.9182             & 0.6496        & 0.6271        & 0.6862            \\
\textbf{0.3}                              & 0.7065          & 0.6303        & 0.6356        & 0.9301         & 0.9301             & 0.6827        & 0.6621        & 0.6862            \\
\textbf{0.4}                              & 0.7462          & 0.6503        & 0.6758        & 0.9308         & 0.9401             & 0.7224        & 0.7077        & 0.6862            \\
\textbf{0.5}                              & 0.7662          & 0.6701        & 0.7118        & 0.9427         & 0.9482             & 0.7531        & 0.7212        & 0.6862 \\          
\midrule
\end{tabular}              
	}
 \vspace{-0.1in}
 	\caption{Avg. ROC-AUC of noisy labels at different noisy label qualities (higher the better). Note that security$^*$'s noisy labels are not simulated and thus do not vary.} % title of Table
	\label{table:label_roc} % is used to refer this table in the text
 \vspace{-0.2in}
\end{table}

\subsection{AD Algorithms and Binary Classifiers}
\label{appx:semiad}

We provide a brief description of AD algorithms below (see more details to recent literature \cite{unifying_shallow_deep,han2022adbench}). Since it lacks specialized fully supervised AD methods, we discuss some SOTA binary classifiers below.
\begin{enumerate}

    \item \textbf{Extreme Gradient Boosting Outlier Detection (XGBOD)} \cite{zhao2018xgbod}. XGBOD uses unsupervised outlier detectors to extract representations for the underlying dataset and concatenates the newly generated features to the original feature feature for augmentation. An XGBoost classifier is then applied to the augmented feature space.
    
    \item \textbf{Deviation Networks (DevNet)} \cite{pang2019devnet} uses a prior probability to enforce a statistical deviation score of input instances. 
    
    \item \textbf{Pairwise Relation prediction-based ordinal regression Network (PReNet)} \cite{pang2019deep} is a neural network-based model that defines a two-stream ordinal regression to learn the relation of instance pairs. 

    \item \textbf{Deep Semi-supervised Anomaly Detection (DeepSAD)} \cite{ruff2019deep}. DeepSAD is considered as the SOTA semi-supervised AD method. It improves the early version of unsupervised DeepSVDD \cite{pmlr-v80-ruff18a} by including the supervision to penalize the inverse of the distances of anomaly representation. In this way, anomalies are forced to be in the space which is away from the hypersphere center. 
    
    \item \textbf{Multi-layer Perceptron (MLP)} \cite{rosenblatt1958perceptron}. MLP is a simple feedforward version of the neural network, which uses the binary cross entropy loss to update network parameters.
    
    \item \textbf{Highly Efficient Gradient Boosting Decision Tree (LightGBM)} \cite{ke2017lightgbm} is a gradient boosting framework that uses tree-based learning algorithms with faster training speed, higher efficiency, lower memory usage, and better accuracy.
    
\end{enumerate}
\vspace{-0.1in}
% \section{Additional Analysis of MoE}
% \label{appx:moe_case}
% \input{tables/expert_sim}

\section{More Experiment Settings and Details}
\label{appx:exp}

\subsection{Code and Reproducibility}
See our datasets and code at GitHub repository \url{ https://github.com/microsoft/admoe}. Part of the code is based on PyOD \cite{zhao2019pyod}, ADBench \cite{han2022adbench}, and TOD \cite{zhao2023tod}.

\subsection{Hyperparameter Setting of \method and Baselines}
\label{appx:hp}
For all baselines used in this study, we use the same set of key hyperparameters for a fair comparison. We also use algorithms default hyperparameter (HP) settings in the original paper for unique hyperparameters. 
More specifically, we use the same: (\textit{i}) learning rate=$0.001$; (\textit{ii}) batch size$=256$; and (\textit{iii}) model size (number of trainable parameters $\approx 18,000$). We use the small validation set (5\%) to choose the epoch with the highest validation performance.
See our code at \url{https://tinyurl.com/admoe22} for more details.

% Specific values can be found in Appx.\ref{appendix:algorithms} and our codebase\footnote{\system repo: \url{https://github.com/Minqi824/ADBench}}. 
% It is also acknowledged that it is possible to use a small hold-out data for hyperparameter tuning for semi- and fully-supervised methods \cite{soenen2021effect}, while we do not consider this setting in this work.

\subsection{Results by Average Precision (AP)}
\label{appx:ap}

\begin{table*}[!htp]
\centering
	\footnotesize
	\vspace{-0.1in}
	\centering
\scalebox{0.9}{
\begin{tabular}{l|cccc|ccc|ccc}
\toprule
\textbf{Dataset}   & \textbf{XGBOD} & \textbf{PreNet} & \textbf{DevNet} & \textbf{LGB} & \textbf{MLP} & \makecell{\textbf{\method-}\\ \textbf{MLP}} & \textbf{$\Delta$ Perf.} & \textbf{DeepSAD} & \makecell{\textbf{\method-}\\ \textbf{DeepSAD}} & \textbf{$\Delta$ Perf.} \\
\midrule
\textbf{agnews}    & 0.2588         & 0.4053          & 0.3851          & 0.2517       & 0.2956       & 0.4775           & +61.54\%            & 0.4212           & \textbf{0.5512}  & 30.86\%           \\
\textbf{aloi}      & 0.0675         & 0.0599          & 0.0616          & 0.0564       & 0.0639       & \textbf{0.1713}  & +168.08\%           & 0.1126           & 0.1226           & 8.88\%            \\
\textbf{imdb}      & 0.1043         & 0.1592          & 0.1097          & 0.0912       & 0.1211       & 0.1699           & +40.30\%            & 0.2312           & \textbf{0.3062}  & 32.44\%           \\
\textbf{mnist}     & 0.9131         & 0.8939          & 0.8721          & 0.9105       & 0.8696       & 0.9284           & +6.76\%             & 0.9014           & \textbf{0.9472}  & 5.08\%            \\
\textbf{spamspace} & 0.9543         & 0.9261          & 0.9327          & 0.954        & 0.946        & 0.959            & +1.37\%             & 0.9327           & \textbf{0.9633}  & 3.28\%            \\
\textbf{svhn}      & 0.3468         & 0.346           & 0.2928          & 0.3457       & 0.2919       & 0.5021           & +72.01\%            & 0.3442           & \textbf{0.5545}  & 61.10\%           \\
\textbf{yelp}      & 0.1827         & 0.2862          & 0.1912          & 0.1519       & 0.1889       & 0.3613           & +91.27\%            & 0.295            & \textbf{0.4882}  & 65.49\%           \\
% \midrule
% \textbf{security$^{*}$}  & 0.7479         & 0.7415          & 0.7498          & 0.7335       & 0.7363       & 0.7767          & +5.49\%              & 0.7928           & \textbf{0.8108} & +2.27\%              \\
\midrule
\midrule
\textbf{avg}       & 0.4039         & 0.4395          & 0.4065          & 0.3945       & 0.3967       & 0.5099           & +63.05\%             & 0.4626           & \textbf{0.5619}  & +29.59\%          \\  
\bottomrule
\end{tabular}
}
	\caption{Average precision (AP) comparison between \method-enhanced AD methods and leading AD methods (that can only use one set of labels) at noisy level 0.2. The best performance is highlighted in bold per dataset (row). \method-based methods (\method-MLP and \method-DeepSAD denote \method-enhanced MLP and DeepSAD) outperform all baselines. \method brings on average 63.05\% improvement over the original MLP, and on average 29.59\% improvement over DeepSAD. Note that all the neural-network models use the equivalent numbers of parameters and FLOPS.} % title of Table
	\label{table:ad_results_ap} % is used to refer this table in the text
	\vspace{-0.15in}
\end{table*}

\subsection{More Ablation Results on the Use of MoE and Weak Inputs}
\begin{figure} [!ht]
    \centering
	\subfloat[\texttt{svhn}] {%
			\includegraphics[clip,width=0.5\columnwidth]{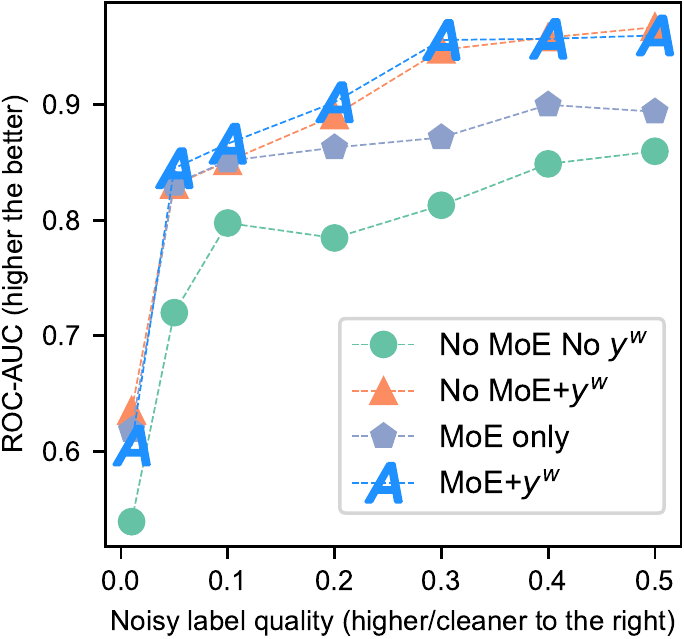}%
		\label{fig:ablation_svhn}
	}
		\subfloat[\texttt{yelp}] {%
			\includegraphics[clip,width=0.5\columnwidth]{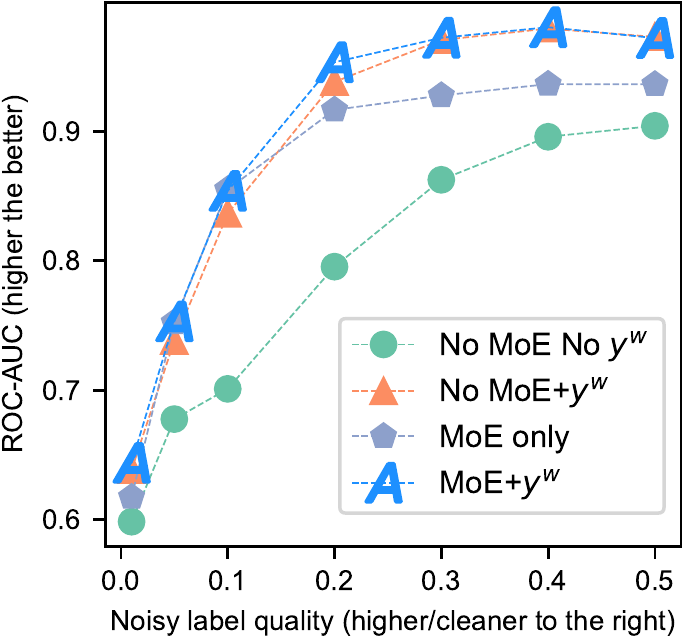}%
		\label{fig:ablation_yelp}
	}
	\vspace{-0.1in}
	\caption{
	Ablation studies on (\textit{i}) the use of \method layer and (\textit{ii}) the noisy labels $\mathbf{y}_w$ as input. \method (\textcolor{dodgerblue}{\textbf{\textit{A}}}) using both techniques shows the best results at (nearly) all settings.
	}
	\label{fig:appx_ablation_moe_input}
	\vspace{-0.2in}
\end{figure}

In addition to the ablation studies in \S 4.4.2, we provide additional results in Fig. \ref{fig:appx_ablation_moe_input} for the effects of using \method and noisy labels as inputs. Similar statements can be made.
First, \method performs the best while using these two techniques jointly in most cases, and significantly better than not using them (\mycircle{SeaGreen}). Second, \method helps the most when labels are noisier (to the left of the x-axis) with an avg. of 2\% improvement over only using noisy labels as input (\mytriangle{orange}). As expected, its impact is reduced with less noisy labels (i.e., closer to the ground truth): 
% when noisy labels are all high quality 
in that case, noisy labels are more similar to each other and specialization with \method is less useful.

\subsection{Additional Results for Effect of Number of Experts ($m$) and top-$k$ Gating.}
\label{appx:ablation_number_experts}
\begin{figure} [!htp]
    \centering
% 	\subfloat[\texttt{agnews}] {%
% 			\includegraphics[clip,width=0.49\columnwidth]{figs/expert_agnews_0.pdf}%
% 		\label{fig:expert_agnews}
% 	}
% % 	\hspace{0.02in}
% 	\subfloat[\texttt{imdb}]{%
% 		\includegraphics[clip,width=0.49\columnwidth]{figs/expert_imdb.pdf}%
% 	\label{fig:expert_imdb}
% 	}
	
% 	\subfloat[\texttt{mnist}] {%
% 			\includegraphics[clip,width=0.49\columnwidth]{figs/expert_27_mnist.pdf}%
% 		\label{fig:expert_mnist}
% 	}
% % 	\hspace{0.02in}
% 	\subfloat[\texttt{spamspace}] {%
% 		\includegraphics[clip,width=0.49\columnwidth]{figs/expert_12_SpamBase.pdf}%
% 		\label{fig:expert_spamspace}
% 	}
	
	\subfloat[\texttt{svhn}] {%
		\includegraphics[clip,width=0.49\columnwidth]{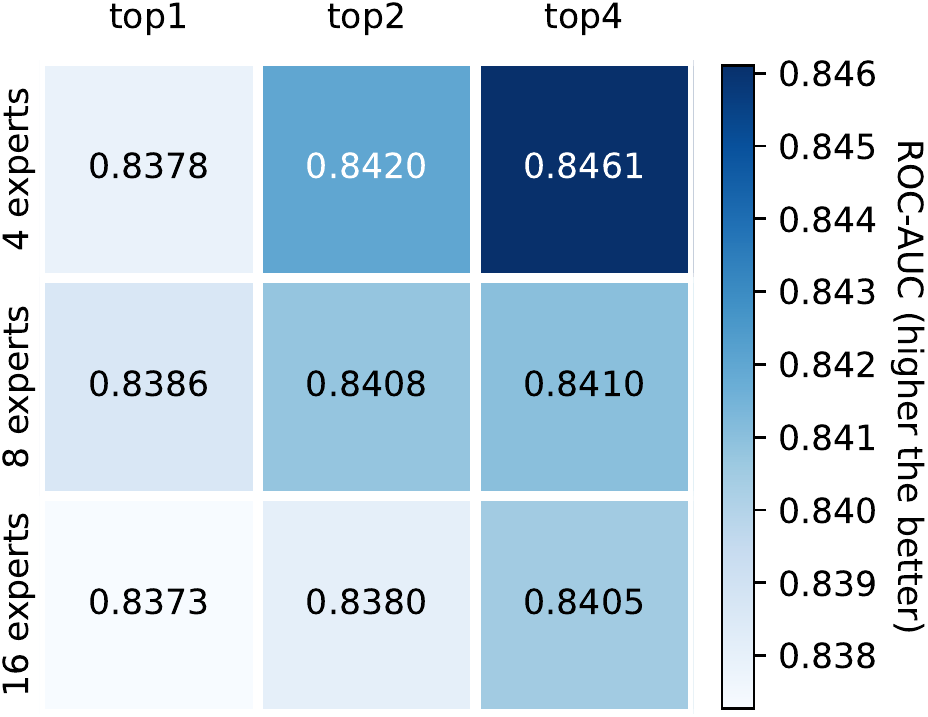}%
		\label{fig:expert_svhn}
	}
	\subfloat[\texttt{yelp}] {%
		\includegraphics[clip,width=0.49\columnwidth]{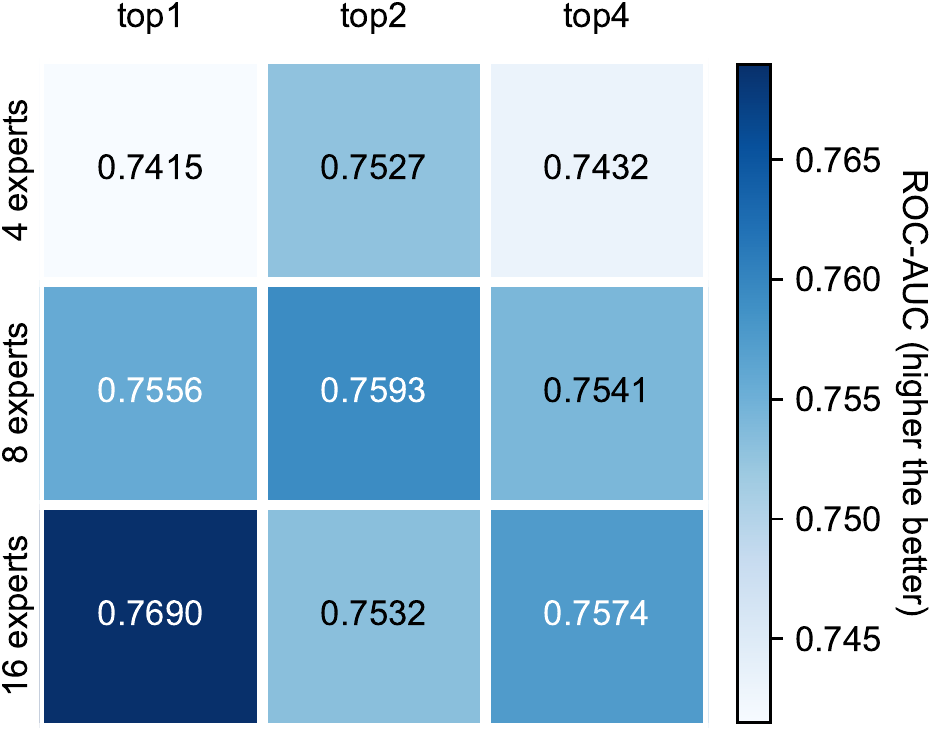}%
		\label{fig:expert_yelp}
	}
	\vspace{-0.1in}
	\caption{
	Additional ablation results on key hyperparameters in \method: (x-axis) the number of experts and (y-axis) top-$k$ experts to activate. We show the results at noisy level 0.05, and find 
 the best setting is data-dependent.
	}
	\vspace{-0.1in}
	\label{fig:ablation_num_experts_appx}
\end{figure}

Fig. \ref{fig:ablation_num_experts_appx} provides additional results for \S 4.4.3 with consistent observations.

\subsection{Additional Results for Performance on Varying Percentage of Clean labels}
\begin{figure} [!htp]
    \centering
	\subfloat[\texttt{agnews}] {%
			\includegraphics[clip,width=0.49\columnwidth]{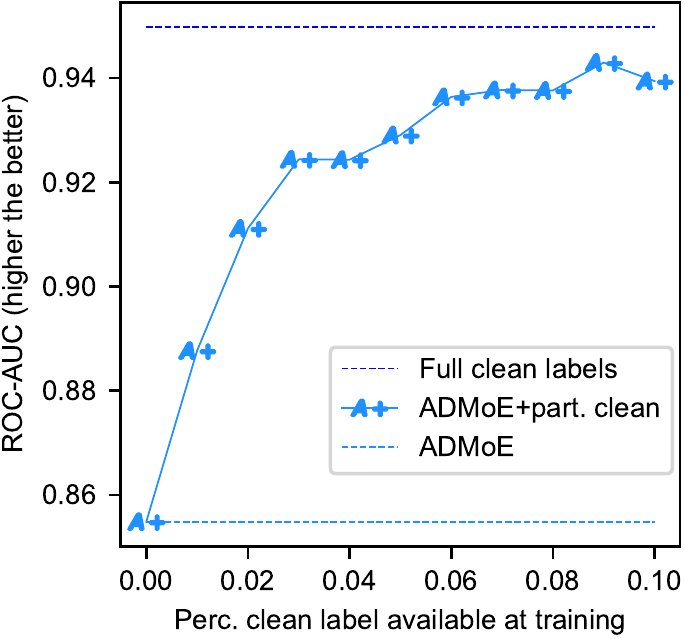}%
		\label{fig:clean_agnews}
	}
% 	\hspace{0.02in}
	\subfloat[\texttt{imdb}]{%
		\includegraphics[clip,width=0.49\columnwidth]{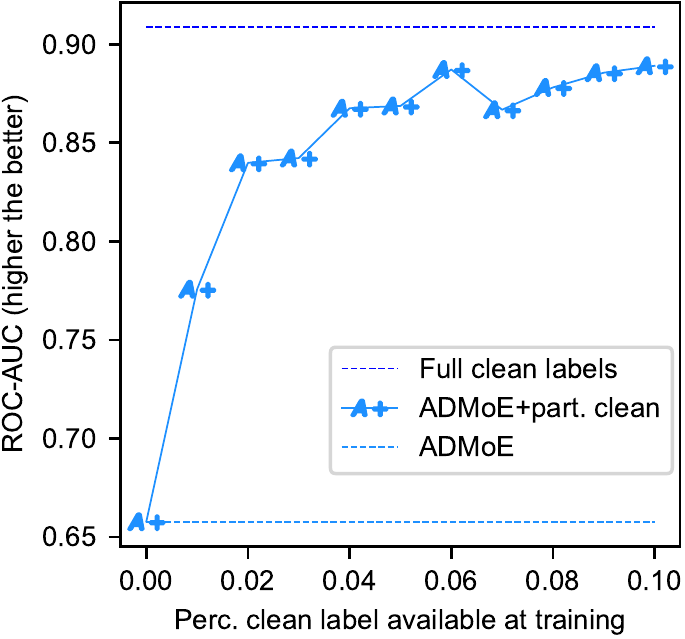}%
	\label{fig:clean_imdb}
	}
	
	\subfloat[\texttt{mnist}] {%
			\includegraphics[clip,width=0.49\columnwidth]{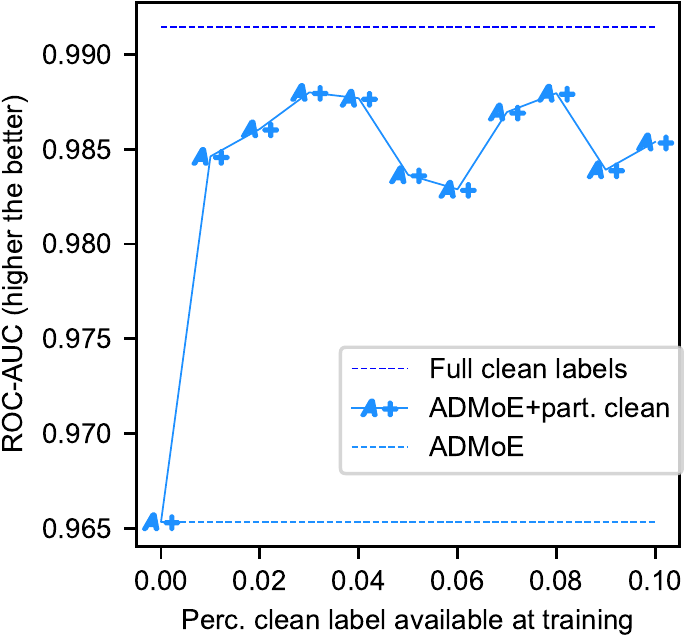}%
		\label{fig:clean_mnist}
	}
	\subfloat[\texttt{yelp}] {%
		\includegraphics[clip,width=0.49\columnwidth]{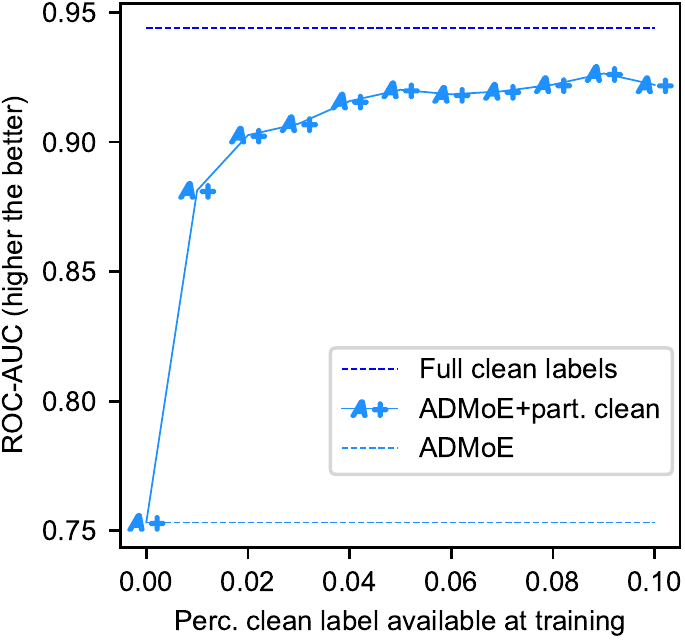}%
		\label{fig:clean_yelp}
	}
	\vspace{-0.1in}
	\caption{
	Additional analysis of integrating varying percentages (from 1\% to 10\%) of \textit{additional} clean labels in \method. The results show that \method efficiently leverages the clean labels with increasing performance.
	}
	\vspace{-0.1in}
	\label{fig:ablation_clean_appx}
\end{figure}

Fig. \ref{fig:ablation_clean_appx} provides additional results for \S 4.4.4 with consistent observations.

\label{sec:appendix}

\end{document}